%% file: main.tex
\documentclass[11pt]{article}

\usepackage[margin=1in]{geometry}
\usepackage{times}
\usepackage[round,authoryear]{natbib}

\input{math_commands.tex}

\usepackage[hidelinks]{hyperref}
\usepackage{url}
\usepackage{amsmath, amssymb}
\usepackage{booktabs}
\usepackage[table]{xcolor}
\usepackage{pifont}
\usepackage{enumitem}
\usepackage{graphicx}
\usepackage{subcaption}
\usepackage{wrapfig}

\newcommand{\xmark}{\ding{55}}
\newcommand{\cmark}{\ding{51}}

\title{
StairVLA: Stage-Aware Hierarchical Action Generation for
Vision-Language-Action Models
}

\author{
Shangyuan Yuan$^{1}$,
Xinda Qi$^{1,2}$,
Yujiang Pu$^{1}$,
Wenliang Guo$^{1}$,
Xiaobo Tan$^{1}$
\\[1mm]
$^{1}$Michigan State University
\qquad
$^{2}$Ant Technology U.S., Inc.
\\
\texttt{\{yuanshan, qixinda, puyujian, guowenli, xbtan\}@msu.edu}
}

\date{}

\begin{document}

\maketitle

\input{sec/0_abstract}
\input{sec/1_introduction_v3}
\input{sec/2_related_work}
\input{sec/3_method_v2}
\input{sec/4_experiments}
\input{sec/5_conclusion}
\input{sec/6_statement}

\bibliographystyle{plainnat}
\bibliography{ref}

\appendix
\section*{Appendix}
This appendix is organized as follows:
\begin{itemize}[leftmargin=0.5cm]
\vspace{-2mm}
    \item \textbf{Appendix \ref{section:a1}} provides additional implementation details.
    \item \textbf{Appendix \ref{section:a2}} presents additional simulation results and analyses.
    \item \textbf{Appendix \ref{section:a3}} provides detailed real-world experimental settings, task definitions, evaluation protocols, and additional analyses.
\end{itemize}
\input{sec/a1_details}
\input{sec/a2_experiments_all}
\input{sec/a3_realworld}

\end{document}

%% file: math_commands.tex
\usepackage{amsmath,amsfonts,bm}

\def\eqref#1{equation~\ref{#1}}

\def\1{\bm{1}}

\DeclareMathAlphabet{\mathsfit}{\encodingdefault}{\sfdefault}{m}{sl}
\SetMathAlphabet{\mathsfit}{bold}{\encodingdefault}{\sfdefault}{bx}{n}



%% file: sec/0_abstract.tex
\begin{abstract}
Vision-language-action (VLA) models increasingly rely on diffusion- or flow-matching-based action heads to generate continuous robot actions. These action heads typically process the denoising trajectory in a largely uniform manner. However, we observe that the conditioning focus naturally shifts across denoising stages: early stages combine language instructions and visual observations to establish a coarse action trajectory, whereas later stages place greater emphasis on current visual observations for action alignment. Based on this insight, we introduce StairVLA, a stage-aware hierarchical action generation framework that uses partially denoised actions as a natural interface between coarse long-horizon action generation and local refinement. A high-level VLA performs early denoising to produce a reusable long-horizon partially denoised action trajectory, while a lightweight refiner operates at a higher frequency to refine local action chunks using the latest observations. This design amortizes expensive high-level VLA computation while preserving frequent closed-loop correction. On LIBERO, our GR00T-style instantiation improves average success from 96.5\% to 97.8\% while reducing amortized inference latency from 115.0 ms to 44.2 ms per action chunk. More broadly, across two VLA backbones, simulation benchmarks, and real-robot tasks, StairVLA consistently reduces inference cost while maintaining strong task performance.
\textbf{Project page:} \url{https://dicomsky.github.io/projects/stairvla}
\end{abstract}

%% file: sec/1_introduction_v3.tex
\section{Introduction}

Vision-language-action (VLA) models have demonstrated strong performance across a wide range of robot manipulation tasks \citep{kim2024openvla}, combining vision-language understanding with generative action heads to translate multimodal context into continuous robot actions \citep{black2024pi0,bjorck2025gr00t}. Specifically, these diffusion- or flow-matching-based action heads are conditioned on language instructions, visual observations, and robot states to progressively transform noise into executable action trajectories \citep{chi2025diffusion}. Despite this progressive generation process, multimodal conditioning is typically applied throughout the entire denoising trajectory, leaving open whether different types of conditioning information should always be processed together during action generation.



Intuitively, language instructions typically specify task intent that remains stable over many control steps, whereas visual observations and robot states continuously change during execution. Task-level information therefore need not be processed as frequently as the feedback required for precise closed-loop control. Existing hierarchical VLAs exploit this temporal asymmetry by processing high-level information at a lower frequency and communicating it to higher-frequency controllers through subgoals, waypoints, macro-actions, or learned representations \citep{Fang2024MOKAOR,li2025hamster,wei2026libra, du2026cf}. Such hierarchies are typically introduced through externally designed representations or model components. This leads to our central question: \emph{can such a hierarchy arise directly from the generative action process itself?}

\begin{figure*}[t]
    \centering
    \includegraphics[width=\textwidth]{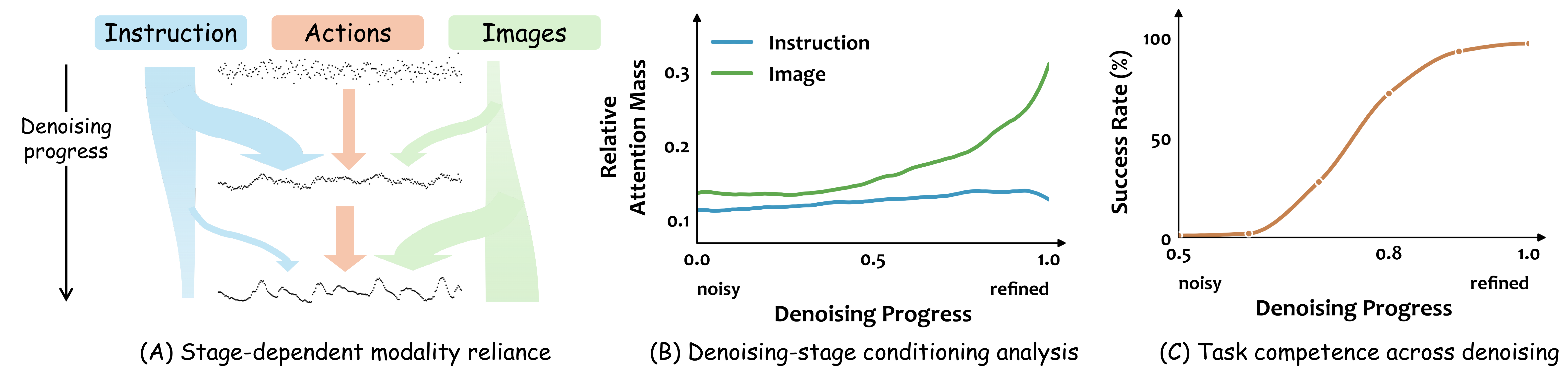}
    \caption{
    Stage dependence in generative VLA action formation.
    \textbf{(A)} Conceptual illustration of the evolving conditioning focus across denoising stages, from joint instruction--observation conditioning toward stronger visual alignment.
    \textbf{(B)} Relative attention to instruction and image tokens across denoising stages in $\pi_0$ on LIBERO, showing an increasing emphasis on visual observations during later refinement.
    \textbf{(C)} Task success of partially denoised actions generated by GR00T on LIBERO, showing that substantial task competence emerges before denoising is complete.
    }
    \label{fig:motivation_hierarchical_generation}
    \vspace{-0.4cm}
\end{figure*}

In practice, we observe that the denoising trajectory itself already exhibits such a structure. As illustrated in Fig.~\ref{fig:motivation_hierarchical_generation}, substantial action structure emerges well before denoising is complete, while the interaction between evolving action and multimodal conditioning continues to change throughout denoising stages. Intermediate actions are therefore not merely noisy versions of the final prediction: they already encode meaningful structure, even as subsequent denoising continues to reshape them toward execution. This reveals a natural stage dependence in generative VLA action formation, with different denoising stages serving distinct functional roles.
This behavior is also consistent with the coarse-to-fine progression commonly observed in generative denoising \citep{balaji2022ediffi,falck2025fourierspaceperspectivediffusion}.

This observation suggests a direct way to exploit the hierarchy already present in denoising: expensive task conditioning can be concentrated in less frequent early-stage generation, while later stages perform frequent local refinement using up-to-date observations. However, this separation raises a key challenge: how can the refiner preserve task information without repeatedly processing the full instruction?
Our key insight is that partially denoised actions themselves can bridge the two stages. By the end of early denoising, they already encode task-relevant action structure while remaining open to further refinement. The low-level refiner can therefore operate directly on these intermediate actions together with the latest observation, preserving high-level task information without repeatedly processing the full instruction.

Based on this insight, we introduce \textbf{StairVLA}, a stage-aware hierarchical action generation framework. A high-level VLA performs the early denoising stages over a long-horizon action trajectory and outputs partially denoised actions that are cached across multiple control steps. For each local chunk, a lightweight refiner takes the corresponding local action segment together with the latest observation and cached high-level context, and completes the remaining refinement before execution. This allows expensive task-conditioned generation to be reused across multiple control steps while keeping local actions responsive to the evolving environment. Importantly, the two stages communicate directly through intermediate states of the generative action trajectory, rather than through an externally designed action abstraction.


We evaluate our approach across two VLA backbones, simulation benchmarks, and real-robot manipulation tasks. On LIBERO, our GR00T-style instantiation improves average success from 96.5\% to 97.8\% while reducing amortized inference latency from 115.0 ms to 44.2 ms per action chunk, with consistent performance-efficiency gains on a $\pi$-style VLA backbone. Although performance gradually degrades as the reuse horizon increases, further analysis shows that StairVLA maintains strong performance over longer high-level reuse horizons than prior hierarchical VLA methods.

In summary, our contributions are threefold:
\begin{itemize}[leftmargin=*, nosep]

    \item We reveal stage dependence in generative VLA action formation, showing that task-relevant action structure emerges progressively while the role of multimodal conditioning evolves throughout denoising.

    \item We introduce StairVLA, a stage-aware hierarchical action generation framework that uses partially denoised actions as a natural interface between coarse long-horizon action generation and local observation-conditioned refinement.

    \item We validate the proposed design across two generative VLA backbones, simulation benchmarks, and real-robot tasks, demonstrating substantially reduced inference cost while maintaining strong task performance.

\end{itemize}

\begin{figure*}[t]
    \centering
    \includegraphics[width=\textwidth]{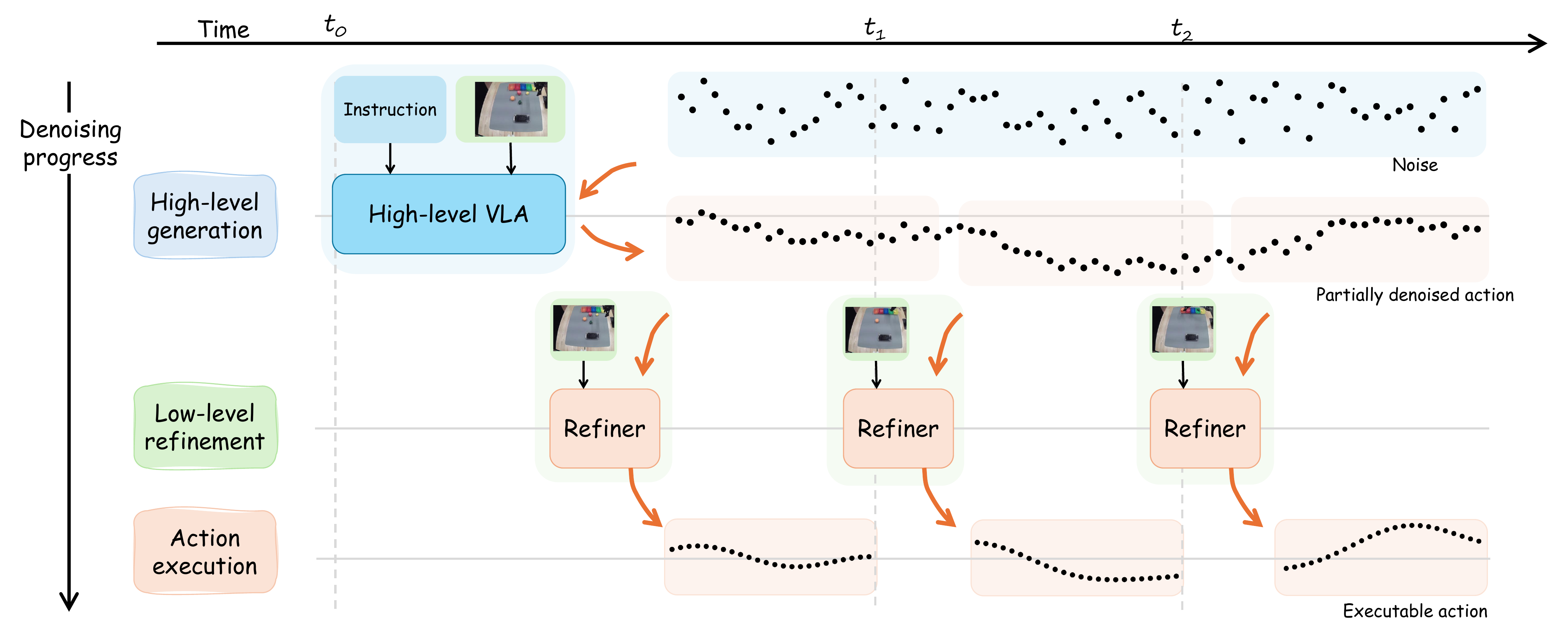}
    \caption{
    Overall framework of StairVLA. The high-level VLA performs early denoising over a long-horizon trajectory and caches the partially denoised trajectory for reuse across multiple steps. For each local chunk, the refiner uses the corresponding segment and the latest observation to complete refinement. The top, middle, and bottom sequences denote the initial noise, partially denoised trajectory, and executable action chunk, respectively.
    }
    \label{fig:method_comparison}
    \vspace{-0.2cm}
\end{figure*}


%% file: sec/2_related_work.tex
\section{Related work}
\label{sec:related_work}

\paragraph{Action Chunking and Unified VLA Models.}
Unified VLA models typically map language instructions, visual observations, and robot states into action sequences within a single policy. Recent VLA models further adopt generative action heads based on diffusion or flow matching to produce continuous actions~\citep{black2024pi0,bjorck2025gr00t,chi2025diffusion}. These methods commonly use action chunking, predicting a sequence of future actions rather than a single-step action to improve temporal consistency and reduce policy invocation frequency~\citep{Zhao-RSS-23,chi2025diffusion}. However, action chunking introduces an inherent horizon trade-off: short chunks require frequent invocation of computationally expensive VLMs, while long chunks weaken closed-loop feedback and real-time correction during execution~\citep{zhang2025action}. Therefore, processing different information and generating actions at a single temporal scale may be unnecessarily restrictive.

\paragraph{Hierarchical VLA Models.}
Recent works increasingly adopt hierarchical structures to bridge high-level task understanding and low-level robot control. Earlier methods use explicit intermediate goals to guide low-level policies, such as subgoals, waypoints, keypoints, or language sub-instructions~\citep{hu2023look,Fang2024MOKAOR,li2025hamster,shi2025hi}. More recent works further integrate this hierarchy into the model architecture and explore different high-to-low-level interfaces, including VLM tokens, learned features, discrete macro-actions, and coarse action representations~\citep{cui2025openhelix,guo2026brain,wei2026libra,wu2026coarse}. Other dual-system approaches focus more explicitly on temporal organization, using event-driven high-level updates or allowing different levels to operate at different frequencies~\citep{chen2026streamvla,sun2026unifs}. Despite their different forms, these methods highlight a common design question in hierarchical VLA: how should high-level task intent be transferred to low-level action generation? Rather than using manually designed or learned interfaces, we directly leverage intermediate states formed during action generation as a natural high-to-low-level interface.

\paragraph{Generative Action Models and Intermediate Denoising States.}
Diffusion- and flow-based generation progressively transforms noise into executable actions. In both generative modeling and robot policies, prior works assign different denoising experts to different sampling stages to better accommodate stage-specific generation requirements~\citep{balaji2022ediffi,zhang2024improving,reuss2025efficient,du2026cf,niu2026t}. Another line of work reuses partially denoised trajectories or constructs better warm starts to reduce redundant sampling and control latency~\citep{hoeg2025fast,chen2025responsive,li2026step}. Our method further uses partially denoised actions to transfer task intent across stages, enabling high-level coarse action generation and local refinement to use different conditioning and operate at different frequencies.

%% file: sec/3_method_v2.tex
\section{Method}
\label{sec:method}

\subsection{Overview}

As illustrated in Fig.~\ref{fig:method_comparison}, we decompose VLA action generation into two stages that operate at different frequencies. The high-level VLA is invoked at a lower frequency to perform early denoising and produce a long-horizon partially denoised action trajectory, which is reused across multiple local action chunks. The low-level refiner runs at a higher frequency, taking the corresponding local action segment and the latest visual observation to produce an executable action chunk. This design reduces repeated VLM computation while preserving closed-loop correction with updated observations.

Our framework is designed to be applicable to flow-matching-based generative VLAs and is trained in two stages. The high-level VLA retains the original VLA architecture with an extended action horizon. The lightweight refiner is the main additional module and follows the design of the VLA action head. It starts from partially denoised actions produced by the trained high-level VLA and performs local refinement using current observations. The high-level generation, refiner architecture and training, and multi-rate inference procedures are detailed in the following subsections.

\subsection{High-Level Intent Trajectory Generation}
\label{sec:high_level_generation}

The high-level VLA retains the original generative architecture and training objective, while extending the action horizon to $H$. Taking a flow-matching VLA as an example, training constructs a continuous path between Gaussian noise $\boldsymbol{\epsilon}\sim\mathcal{N}(\mathbf{0},\mathbf{I})$ and the ground-truth $\mathbf{a}^{\mathrm{gt}}$:
\begin{equation}
\mathbf{a}^{\gamma}
=
(1-\gamma)\boldsymbol{\epsilon}
+
\gamma\mathbf{a}^{\mathrm{gt}},
\qquad
\gamma\in[0,1].
\label{eq:flow_path}
\end{equation}
The corresponding target velocity is
\begin{equation}
\mathbf{v}^{*}
=
\mathbf{a}^{\mathrm{gt}}
-
\boldsymbol{\epsilon}.
\label{eq:flow_velocity}
\end{equation}
Given the intermediate action state $\mathbf{a}^{\gamma}$, denoising progress $\gamma$, and VLM conditioning feature $\mathbf{z}_t$, the action head $v_\theta$ is trained to predict this velocity:
\begin{equation}
\mathcal{L}_{\mathrm{FM}}
=
\mathbb{E}
\left[
\left\|
v_\theta
\left(
\mathbf{a}^{\gamma},
\gamma,
\mathbf{z}_t
\right)
-
\mathbf{v}^{*}
\right\|_2^2
\right].
\label{eq:flow_matching_loss}
\end{equation}

At inference time, action generation starts from random noise,
$\mathbf{a}^{\gamma_0}=\boldsymbol{\epsilon}$, and iteratively updates the action state by repeatedly predicting the velocity:
\begin{equation}
\mathbf{a}^{\gamma_{n+1}}
=
\mathbf{a}^{\gamma_n}
+
\left(
\gamma_{n+1}-\gamma_n
\right)
v_\theta
\left(
\mathbf{a}^{\gamma_n},
\gamma_n,
\mathbf{z}_t
\right).
\label{eq:flow_inference}
\end{equation}
A standard VLA continues this process until $\gamma_N=1$ to obtain the final executable action trajectory. In contrast, we terminate the high-level generation at $\gamma_N=\alpha<1$ and use the resulting partially denoised trajectory $\mathbf{a}^{\alpha}_{t:t+H-1}$ as the high-level output. This long-horizon trajectory is then cached and serves as the starting point for subsequent low-level refinement.

\begin{figure}[t]
    \centering
    \includegraphics[width=\linewidth]{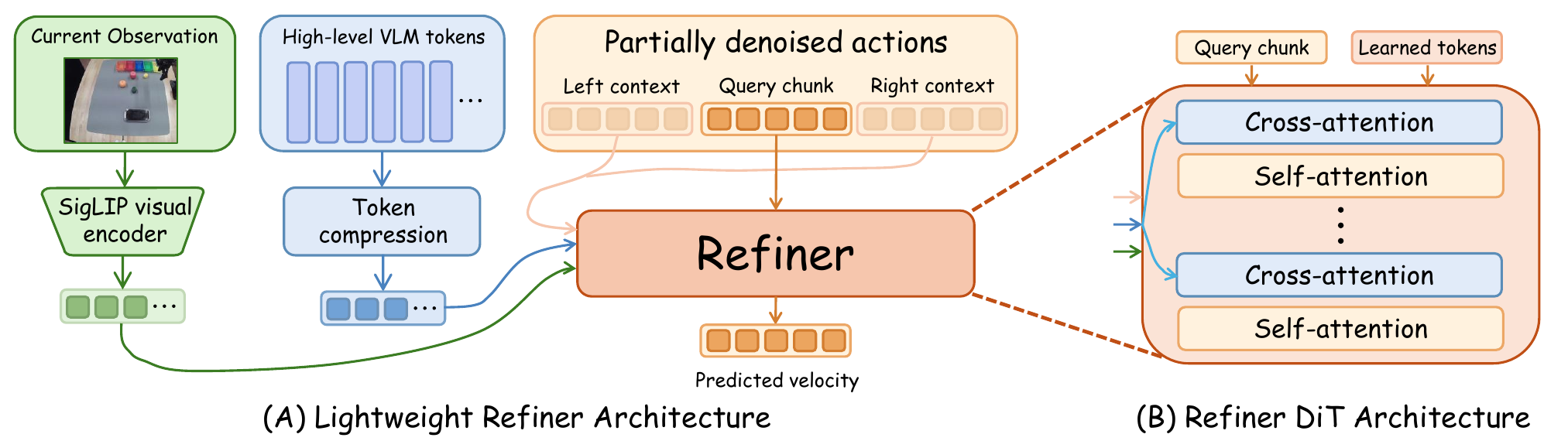}
    \caption{
    Overview of the low-level refiner.
    \textbf{(A)} The refiner uses the current observation, compressed high-level VLM tokens, and partially denoised action context to refine the query chunk.
    \textbf{(B)} The refiner adopts a GR00T-style DiT with interleaved cross-attention and self-attention for conditional fusion and action modeling.
    }
    \label{fig:method_overview}
\end{figure}

\subsection{Observation-Conditioned Low-Level Refinement}
\label{sec:low_level_refinement}

As illustrated in Figure~\ref{fig:method_overview}, we introduce a lightweight refiner to further refine the partially denoised local action segments produced by the high-level VLA. The refiner uses a SigLIP-B/16 visual encoder to process the latest observation, which is jointly trained. Meanwhile, the hidden states from the last layer of the high-level VLM are compressed into a fixed number of tokens through adaptive average pooling, which are cached and reused across multiple refinement cycles.

As illustrated on the right side of Figure~\ref{fig:method_overview}, the refiner follows the overall architecture of the GR00T action head~\citep{bjorck2025gr00t} and takes two types of inputs: action-side tokens and condition-side tokens. The action side includes the robot state token, learnable future tokens, and the current query action chunk, while the condition side consists of visual tokens, compressed VLM tokens, and temporal action context extracted from the cached high-level trajectory. Self-attention operates only on the action-side tokens to model dependencies within the action representation, whereas cross-attention uses the action-side tokens as queries to incorporate information from the condition-side tokens. These two attention layers are interleaved throughout the DiT, alternating between action modeling and conditional fusion.

\begin{wrapfigure}{r}{0.45\textwidth}
    \centering
    \includegraphics[width=\linewidth]{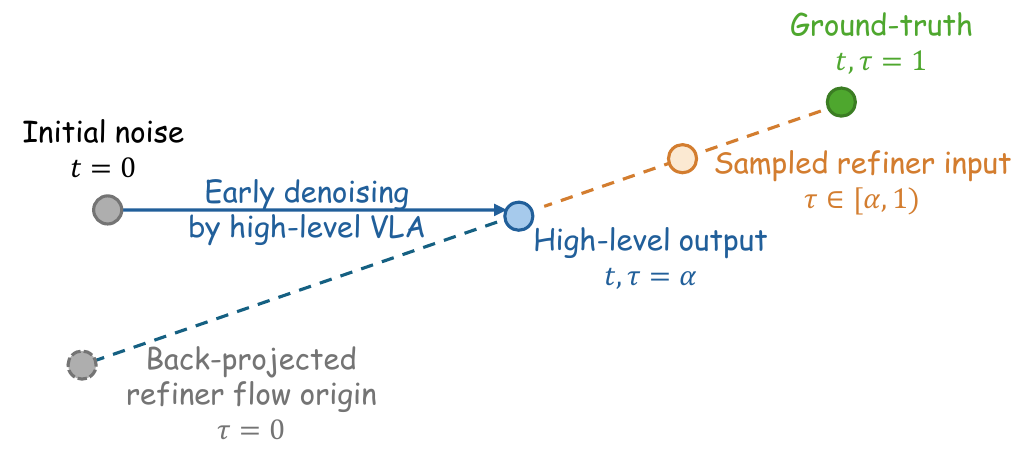}
    \caption{
    Refiner training path construction.
    After early denoising by the high-level VLA, the resulting partially denoised action is
    used to construct the refinement path toward the ground-truth action, from which both
    refiner inputs and targets are generated.
    }
    \label{fig:refiner_training_path}
\end{wrapfigure}

The refiner training does not directly follow the original flow-matching path of the high-level VLA. Due to prediction errors, its partially denoised actions may not lie exactly on the ideal generation path defined by the denoising origin and the ground-truth. Therefore, as illustrated in Figure~\ref{fig:refiner_training_path}, for each local query chunk, we reconstruct a refinement path toward the ground-truth. 

For clarity, we omit the local chunk indices in this subsection. We treat the high-level output $\mathbf{a}^{\alpha}$ as the state of a reconstructed flow at time $\alpha$ and back-project its virtual origin as
\begin{equation}
\boldsymbol{\epsilon}_{\mathrm{ref}}
=
\frac{
\mathbf{a}^{\alpha}
-
\alpha \mathbf{a}^{\mathrm{gt}}
}{
1-\alpha
}.
\label{eq:refiner_origin}
\end{equation}
The corresponding target velocity is
\begin{equation}
\mathbf{v}_{\mathrm{ref}}
=
\mathbf{a}^{\mathrm{gt}}
-
\boldsymbol{\epsilon}_{\mathrm{ref}}
=
\frac{
\mathbf{a}^{\mathrm{gt}}
-
\mathbf{a}^{\alpha}
}{
1-\alpha
}.
\label{eq:refiner_velocity}
\end{equation}

During training, we sample $\tau\sim\mathcal{U}(\alpha,1)$ and construct the intermediate refiner input along the reconstructed path:
\begin{equation}
\mathbf{a}_{\mathrm{ref}}^{\tau}
=
(1-\tau)\boldsymbol{\epsilon}_{\mathrm{ref}}
+
\tau\mathbf{a}^{\mathrm{gt}}.
\label{eq:refiner_path}
\end{equation}
The refiner $v_{\phi}$ is optimized to predict the reconstructed velocity:
\begin{equation}
\mathcal{L}_{\mathrm{ref}}
=
\mathbb{E}
\left[
\left\|
v_{\phi}
\left(
\mathbf{a}_{\mathrm{ref}}^{\tau},
\tau,
\mathbf{c}_{\mathrm{ref}}
\right)
-
\mathbf{v}_{\mathrm{ref}}
\right\|_2^2
\right],
\label{eq:refiner_loss}
\end{equation}
where $\mathbf{c}_{\mathrm{ref}}$ denotes the condition-side inputs described above.

\subsection{Multi-Rate Inference and Horizon Reuse}
\label{sec:multirate_inference}

During inference, each high-level VLA invocation generates and caches a partially denoised trajectory of horizon $H$, which is reused across the following $M$ low-level refinement cycles. At each cycle, the low-level refiner extracts the corresponding partially denoised segment of length $h$ from the cached trajectory and further refines it.

Let $T_{\mathrm{high}}$ and $T_{\mathrm{ref}}$ denote the inference time of one high-level invocation and one low-level refinement, respectively. The average inference cost per local action chunk is
\begin{equation}
\overline{T}(M)
=
T_{\mathrm{ref}}
+
\frac{T_{\mathrm{high}}}{M}.
\label{eq:amortized_inference_cost}
\end{equation}

As $M$ increases, the high-level VLA cost is amortized over more local action chunks. However, a larger $M$ also reuses the same high-level trajectory for a longer horizon, which may reduce its consistency with the latest environment state. Therefore, $M$ controls the trade-off between computational efficiency and the freshness of high-level information. We evaluate this trade-off under different reuse horizons in Section~\ref{sec:experiments}.

%% file: sec/4_experiments.tex
\section{Experiments}
\label{sec:experiments}

We evaluate StairVLA in terms of task performance, inference efficiency and cross-backbone generality, followed by analyses of stage-dependent denoising behavior, key design choices and real-world deployment.

\subsection{Model and Training Setup}
\paragraph{Model Configuration.}
We implement our method based on StarVLA~\citep{ye2026starvla}. By default, the high-level policy uses Qwen3-VL-4B-Instruct as the vision-language backbone with a GR00T-style flow-matching action head, and predicts an action horizon of $H=32$. The low-level refiner outputs action chunks of length $h=5$ and uses a trainable SigLIP-B/16 visual encoder. In addition, the refiner uses $k=5$ high-level actions before and after the query chunk as temporal context.

\paragraph{Training Setup.}
We adopt a two-stage training strategy. On LIBERO, we first train the high-level VLA for 30k steps and then freeze the high-level policy to train the low-level refiner for another 30k steps. By default, the refiner is trained using high-level trajectories at denoising progress $\alpha=0.8$. We use AdamW with a global batch size of 128, an initial learning rate of $5\times10^{-5}$, cosine learning-rate decay, 1k warmup steps, and a weight decay of 0.01. All models are trained with bf16 mixed precision on 8 NVIDIA RTX A6000 GPUs.

\paragraph{Inference and Latency Measurement.}
Unless otherwise specified, we use high-level trajectories at $\alpha=0.97$ and reuse each trajectory for $M=4$ refinement cycles, with $h=5$ actions per cycle. Latency is measured per executable action chunk and includes the amortized high-level inference cost and low-level refinement cost, excluding environment execution. All measurements are conducted on a single NVIDIA A100 GPU. Additional implementation, training, and preprocessing details are provided in Appendix~\ref{section:a1}.

\begin{table}[t]
\centering
\caption{
Comparison on the LIBERO benchmark. We report success rates (\%) on the four LIBERO suites, the average success rate (\%), and reported inference latency (ms/chunk) when available.
Chunk sizes may vary across methods.
Best and second-best results are shown in \textbf{bold} and \underline{underlined}, respectively.
$^{*}$ Uses additional subtask-level supervision beyond standard action demonstrations and is excluded from best/second-best ranking.
$^{\ddagger}$ StreamVLA reports average wall-clock latency under its gated reasoning schedule.
$^{\S}$ CF-VLA reports action-sampling latency excluding visual-language prefix encoding and KV-cache construction.
}
\label{tab:libero_main}
\resizebox{0.83\textwidth}{!}{
\begin{tabular}{l c c c c c c}
\toprule
\textbf{Method}
& \textbf{Spatial}
& \textbf{Object}
& \textbf{Goal}
& \textbf{Long}
& \textbf{Avg.}
& \textbf{Latency} \\
\midrule

\multicolumn{7}{c}{\textit{Representative VLA Policies}} \\
\midrule

OpenVLA~\citep{kim2024openvla}
& 84.7 & 88.4 & 79.2 & 53.7 & 76.5 & -- \\

\(\pi_0\)~\citep{black2024pi0}
& 96.8 & 98.8 & 95.8 & 85.2 & 94.1 & -- \\

GR00T~\citep{bjorck2025gr00t}
& 94.4 & 97.6 & 93.0 & 90.6 & 93.9 & -- \\

\(\pi_{0.5}\)~\citep{intelligence2025pi05}
& 98.8 & 98.2 & \underline{98.0} & 92.4 & 96.9 & -- \\

\midrule
\multicolumn{7}{c}{\textit{Hierarchical / Coarse-to-Fine VLA Policies}} \\
\midrule

HiMoE-VLA~\citep{du2025himoe}
& 98.2 & 99.4 & \textbf{98.6} & 94.8 & 97.8 & -- \\

StreamVLA~\citep{chen2026streamvla}
& 99.2$^{*}$ & 99.4$^{*}$ & 98.6$^{*}$ & 96.6$^{*}$ & 98.5$^{*}$ & 128.0$^{\ddagger}$ \\

CF-VLA~\citep{du2026cf}
& 98.0 & 99.2 & 96.6 & 92.0 & 96.5 & 7.81$^{\S}$ \\

Coarse-to-Control~\citep{wu2026coarse}
& 98.8 & \textbf{100.0} & 97.8 & 95.0 & \underline{97.9} & -- \\

Libra-VLA~\citep{wei2026libra}
& 98.6 & 99.4 & \underline{98.0} & 92.8 & 97.2 & 122.0 \\

\midrule
\multicolumn{7}{c}{\textit{StarVLA Baselines and Ours}} \\
\midrule

StarVLA-GR00T~\citep{ye2026starvla}
& 97.8 & 98.8 & 97.4 & 92.0
& 96.5
& 115.0 \\

\rowcolor{gray!12}
Ours (GR00T-base)
& 98.0 & \underline{99.6} & 97.8 & \underline{95.8}
& 97.8 {\scriptsize $(+1.3)$}
& 44.2 \\

StarVLA-$\pi$~\citep{ye2026starvla}
& \underline{99.2} & 99.0 & 97.2 & \underline{95.8}
& 97.8
& 183.3 \\

\rowcolor{gray!12}
Ours ($\pi$-base)
& \textbf{99.6} & 99.0 & 96.8 & \textbf{97.4}
& \textbf{98.3} {\scriptsize $(+0.5)$}
& 44.5 \\

\bottomrule
\end{tabular}
}

\vspace{2pt}
\begin{minipage}{0.92\linewidth}
\end{minipage}
\end{table}

\subsection{Main Results}

\paragraph{LIBERO.}
We first evaluate task performance and inference efficiency on LIBERO~\citep{liu2023libero}. We jointly train a single model on the Spatial, Object, Goal, and Long suites, and evaluate each task with 50 rollouts. As shown in Table~\ref{tab:libero_main}, our method achieves strong task performance across both VLA backbones, while also maintaining a low average inference latency. Compared with existing hierarchical and coarse-to-fine approaches, our method provides a competitive trade-off between task success and inference efficiency. We note that some prior methods use additional subtask-level supervision, while latency definitions, measurement scopes, and hardware settings also vary between works. Therefore, the reported efficiency results are intended to reflect overall trends rather than serve as strictly comparable absolute runtimes.

\begin{table*}[t]
\centering
\small
\setlength{\tabcolsep}{4.5pt}
\renewcommand{\arraystretch}{1.12}
\caption{
Results on LIBERO-Plus under zero-shot transfer and supervised fine-tuning.
}
\label{tab:libero_plus}
\resizebox{0.9\textwidth}{!}{
\begin{tabular}{l ccccccc c}
\toprule
\textbf{Method}
& \textbf{Camera}
& \textbf{Robot}
& \textbf{Language}
& \textbf{Light}
& \textbf{Background}
& \textbf{Noise}
& \textbf{Layout}
& \textbf{Avg.} \\
\midrule
\multicolumn{9}{c}{\textit{Zero-Shot Transfer}} \\
\midrule
OpenVLA~\citep{kim2024openvla}
& 0.8 & 3.5 & 23.0 & 8.1 & 34.8 & 15.2 & 28.5 & 15.6 \\
$\pi_0$-FAST~\citep{pertsch2025fast}
& \underline{65.1} & 21.6 & 61.0 & 73.2 & 73.2 & 74.4 & 68.8 & 61.6 \\
OpenVLA-OFT~\citep{kim2025finetuningvisionlanguageactionmodelsoptimizing}
& 56.4 & 31.9 & 79.5 & 88.7 & \underline{93.3} & \underline{75.8} & \underline{74.2} & 69.6 \\
GR00T~\citep{bjorck2025gr00t}
& 34.6 & \underline{50.8} & \underline{85.1} & 86.5 & 86.5 & 63.0 & 73.6 & 66.8 \\
\midrule
Libra-VLA~\citep{wei2026libra}
& \textbf{68.9} & 48.8 & \textbf{92.7} & \textbf{97.9} & \textbf{93.4} & \textbf{86.3} & \textbf{77.5} & \textbf{79.5} \\
\rowcolor{gray!15}
\textbf{Ours}
& 43.7 & \textbf{54.5} & 82.5 & \underline{93.4} & 90.5 & 67.8 & 74.0 & \underline{70.4} \\
\midrule
\multicolumn{9}{c}{\textit{Supervised Fine-Tuning}} \\
\midrule
$\pi_0$~\citep{black2024pi0}
& 79.6 & 21.1 & 72.5 & 84.7 & 86.2 & 68.3 & 69.4 & 67.4 \\
$\pi_{0.5}$~\citep{intelligence2025pi05}
& 70.3 & 41.7 & 81.1 & \textbf{97.3} & \underline{94.6} & 71.8 & \textbf{84.9} & 75.7 \\
OpenVLA-OFT+~\citep{kim2025finetuningvisionlanguageactionmodelsoptimizing}
& 92.8 & 30.3 & \textbf{85.8} & 94.9 & 93.9 & 89.3 & \underline{77.6} & 79.6 \\
\midrule
Libra-VLA~\citep{wei2026libra}
& \underline{94.5} & \underline{41.8} & 83.2 & 95.3 & 94.3 & \underline{93.7} & 75.3 & \underline{82.3} \\
\rowcolor{gray!15}
\textbf{Ours}
& \textbf{95.5} & \textbf{48.4} & \underline{83.7} & \underline{96.8} & \textbf{94.9} & \textbf{95.4} & 77.4 & \textbf{83.7} \\
\bottomrule
\end{tabular}
}
\end{table*}

\paragraph{LIBERO-Plus.}
We further evaluate our method on LIBERO-Plus~\citep{fei2026libero}, which introduces seven types of perturbations to the original LIBERO tasks. Following the standard protocol, we consider both zero-shot transfer and supervised fine-tuning. In the zero-shot setting, the model trained on LIBERO is directly evaluated in the perturbed environments, while in the supervised setting, it is further fine-tuned on the LIBERO-Plus training set using the same hyperparameter configuration as in LIBERO. As shown in Table~\ref{tab:libero_plus}, our method achieves competitive overall performance in the zero-shot setting. However, since the low-level refiner mainly relies on visual observations for local correction, performance drops more noticeably under image-related perturbations such as camera changes. After supervised fine-tuning, performance recovers substantially across these perturbation categories, reaching an average success rate of $83.7\%$ and competitive results on multiple categories. These results suggest that, although it is sensitive to zero-shot visual shifts, it can effectively adapt to new visual conditions through target-distribution fine-tuning.

\paragraph{Across VLA Backbones.}
We apply our method to both GR00T-based and $\pi$-based StarVLA backbones. As shown in Table~\ref{tab:libero_main}, hierarchical refinement improves the average success rate on both, demonstrating that the method is not limited to a specific VLA backbone. Complete results across different reuse horizons for both backbones are provided in Appendix~\ref{sec:backbone_transfer}.

\begin{figure*}[t]
\centering

\begin{minipage}[t]{0.36\textwidth}
    \centering
    \includegraphics[width=\linewidth]{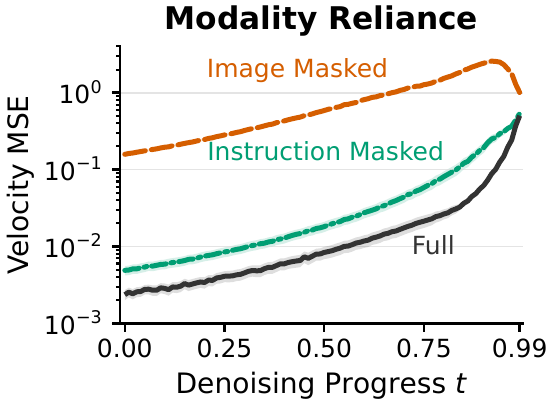}
    \captionof{figure}{
    Conditioning sensitivity across denoising stages on $\pi_0$.
    }
    \label{fig:denoising_reliance}
\end{minipage}
\hfill
\begin{minipage}[t]{0.58\textwidth}
    \centering
    \includegraphics[width=\linewidth]{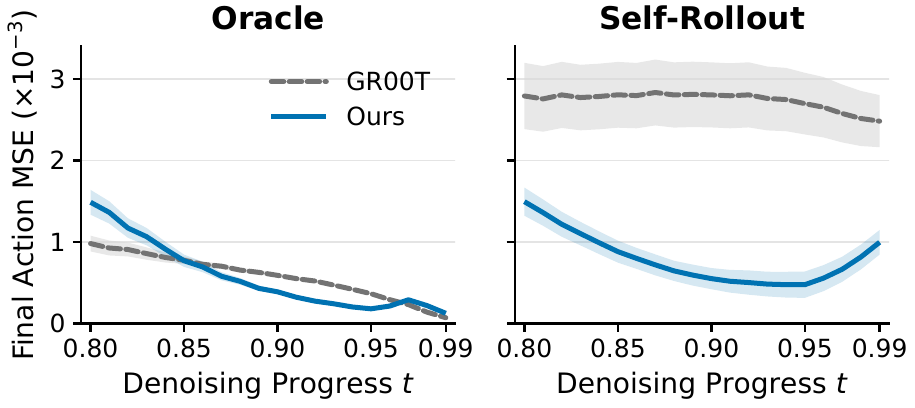}
    \captionof{figure}{
    Final action error of late-stage refinement under oracle and self-rollout settings.
    }
    \label{fig:refinement_comparison}
\end{minipage}

\end{figure*}

\subsection{Analysis of Stage-Aware Action Generation}
\paragraph{Stage-Dependent Denoising Behavior.}
We analyze a trained StarVLA-$\pi$ model using 256 randomly sampled LIBERO examples and measure velocity prediction error at different denoising progress levels. As shown in Fig.~\ref{fig:denoising_reliance}, even without any masking, the error rises sharply near the end of denoising, suggesting that a unified generation mechanism is not equally effective across all stages. Masking the image further causes a larger error increase at later stages, supporting the stronger dependence on visual observations during late-stage refinement. This effect decreases again near completion, possibly because the remaining residuals can be largely corrected through simple local smoothing. These results motivate our stage-wise generation design and observation-conditioned refiner. Additional denoising-stage conditioning analyses are provided in Appendix~\ref{sec:denoising_stage_conditioning}.

\paragraph{Effectiveness of Late-Stage Refinement.}
We randomly sample 1024 LIBERO examples and evaluate two settings: inputs constructed along the training denoising trajectory, and intermediate actions obtained through iterative inference, where prediction errors accumulate. As shown in Fig.~\ref{fig:refinement_comparison}, we measure the error between the final generated actions and the ground truth. Our refiner maintains low and stable error in both settings, while GR00T degrades substantially once accumulated errors are introduced.

\paragraph{Effect of Additional Instruction Conditioning.}
To investigate whether the low-level refiner requires additional instruction information, we augment the baseline refiner with full, unpooled instruction tokens and compare their performance across different inference-time denoising levels. As shown in Fig.~\ref{fig:textinstr_denoising_sweep}, the additional instruction tokens help recover task-relevant information at higher noise levels, but do not further improve the refiner's \emph{best achievable performance}. In contrast, using lighter task conditioning allows the refiner to focus more on observation-conditioned action alignment, leading to a higher performance ceiling.

\subsection{Ablation Studies and Analysis}

\paragraph{Component Ablation.}
As shown in Table~\ref{tab:component_ablation}, the top-only setting, which only increases the high-level prediction horizon, leads to a clear performance drop. Adding the observation-conditioned refiner substantially recovers performance, while temporal action context provides a further improvement, validating both design choices.

\begin{table}[htbp]
\centering
\small
\setlength{\tabcolsep}{4.5pt}
\renewcommand{\arraystretch}{1.12}
\caption{
Ablations on LIBERO.
}
\label{tab:component_ablation}
\resizebox{0.78\linewidth}{!}{
\begin{tabular}{l|cc|cccc|cc}
\toprule
\textbf{Setting}
& \textbf{Refiner}
& \textbf{Action Context}
& \textbf{Spatial}
& \textbf{Object}
& \textbf{Goal}
& \textbf{Long}
& \textbf{Avg.}
& \textbf{$\Delta$} \\
\midrule
StarVLA-GR00T~\citep{ye2026starvla}
& -- & --
& 97.8 & 98.8 & 97.4 & 92.0 & 96.5 & -- \\
\midrule
Ours w/o refiner (Top-only)
& \xmark & --
& 96.8 & 89.4 & 96.8 & 86.4 & 92.4 & -4.1 \\
Ours w/o Action Context
& \cmark & \xmark
& \textbf{99.2} & 98.6 & 95.6 & 95.4 & 97.2 & +0.7 \\
\rowcolor{gray!15}
\textbf{Ours}
& \cmark & \cmark
& 98.0 & \textbf{99.6} & \textbf{97.8} & \textbf{95.8} & \textbf{97.8} & \textbf{+1.3} \\
\bottomrule
\end{tabular}
}
\end{table}

\begin{figure*}[htbp]
\centering

\begin{minipage}[t]{0.34\textwidth}
    \centering
    \includegraphics[width=\linewidth]{
        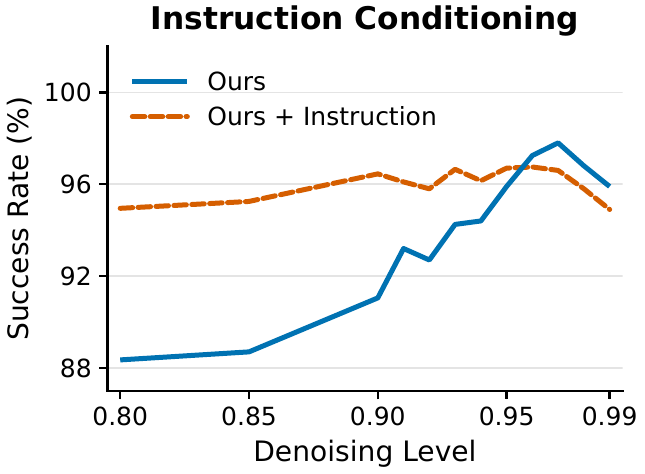
    }
    \captionof{figure}{
    Effect of additional instruction conditioning.
    }
    \label{fig:textinstr_denoising_sweep}
\end{minipage}
\hfill
\begin{minipage}[t]{0.62\textwidth}
    \centering
    \includegraphics[width=\linewidth]{
        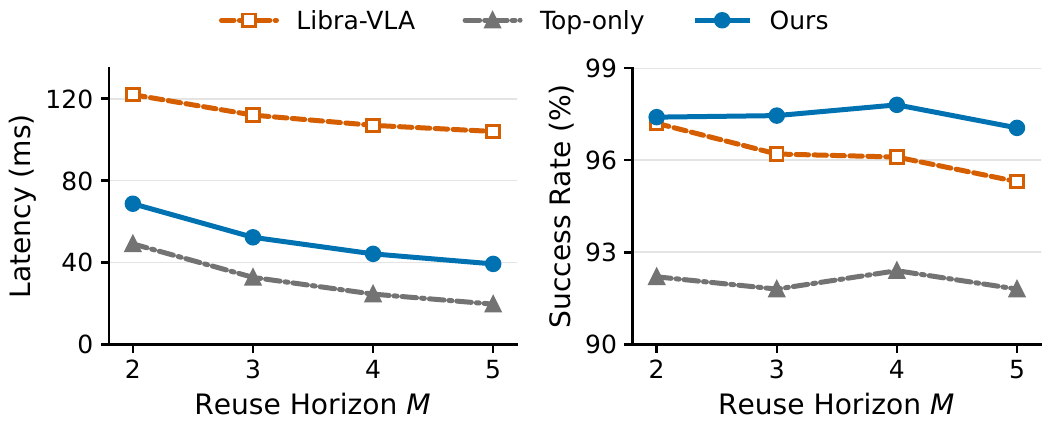
    }
    \captionof{figure}{
    Effect of high-level reuse horizon \(M\) on inference latency and task success.
    }
    \label{fig:reuse_horizon}
\end{minipage}

\end{figure*}

\paragraph{High-Level Trajectory Reuse Horizon.}
We study the effect of the reuse length $M$ on performance and efficiency, where $M$ denotes the number of low-level refinement cycles executed between two high-level VLA updates. As shown in Fig.~\ref{fig:reuse_horizon}, our method maintains strong task performance under longer high-level reuse horizons while substantially reducing inference latency, leading to a more favorable efficiency--performance trade-off. Additional experiments are provided in Appendix~\ref{sec:long_horizon_reuse}.

\begin{figure*}[!t]
\centering

\begin{minipage}[t]{0.34\textwidth}
    \centering
    \includegraphics[width=\linewidth]{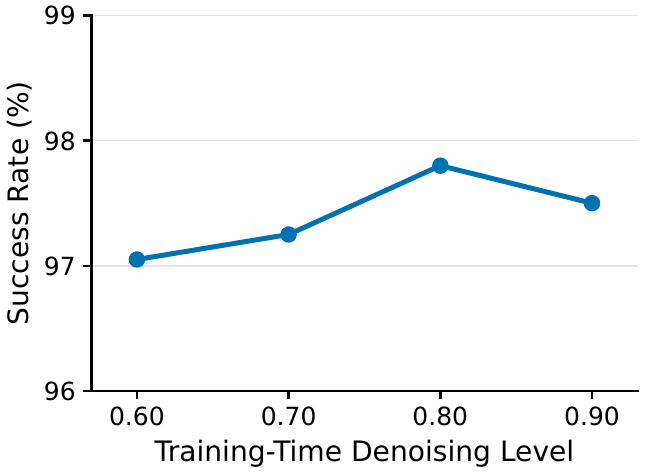}
    \captionof{figure}{
    Effect of the training-time denoising level.
    }
    \label{fig:denoising_level}
\end{minipage}
\hfill
\begin{minipage}[t]{0.62\textwidth}
    \centering
    \includegraphics[width=\linewidth]{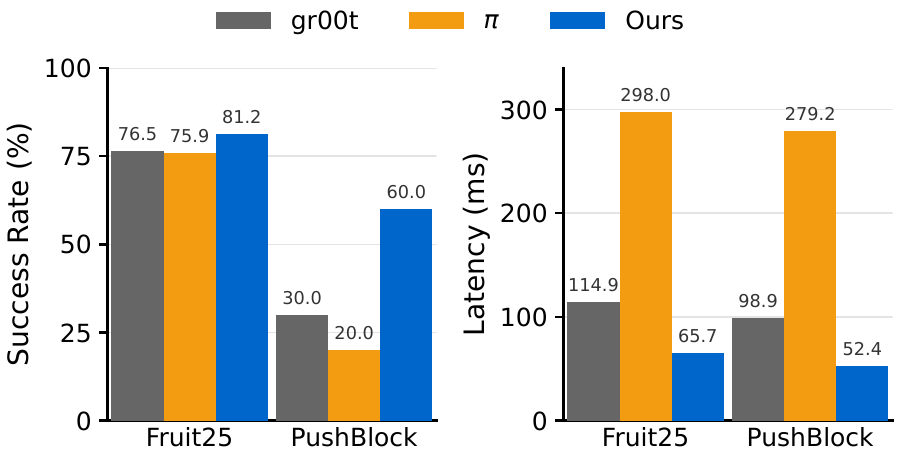}
    \captionof{figure}{
    Real-world results on Fruit25 and PushBlock.
    }
    \label{fig:real_world}
\end{minipage}
\vspace{-0.3cm}
\end{figure*}

\paragraph{Denoising Progress of the Transferred Trajectory.}
We further study the denoising progress of the transferred high-level trajectory during both training and inference. As shown in Fig.~\ref{fig:denoising_level}, the refiner is relatively robust to the training-time transfer point $\alpha$ over the tested range. At inference time, the baseline sweep shown in Fig.~\ref{fig:textinstr_denoising_sweep} indicates that performance improves as the transferred trajectory approaches complete denoising and peaks near $\alpha=0.97$. Notably, we directly use this inference-time setting across other benchmarks and backbone variants without retuning, while still maintaining strong performance. These results suggest that the transfer point around $0.97$ provides a useful balance between task information already encoded in the partially denoised trajectory and the residual space available for low-level refinement.

\subsection{Real-World Experiments}
\label{sec:real_world}

We further evaluate our method on an AgileX PiperX robot in two real-world settings. \textit{Fruit25} includes 24 language-conditioned fruit pick-and-place tasks and one long-horizon multi-object sorting task, while \textit{PushBlock} consists of a single continuous pushing task. We compare against StarVLA-GR00T and StarVLA-$\pi$~\citep{ye2026starvla}, all using Qwen3-VL-2B-Instruct as the vision-language backbone and evaluated on the same NVIDIA RTX A4000 GPU with the same training data and evaluation protocol. As shown in Fig.~\ref{fig:real_world}, the real-world results exhibit performance and efficiency gains consistent with those observed in simulation. Additional details on the real-world platform, datasets, training and evaluation protocols, and per-task results are provided in Appendix~\ref{section:a3}.

%% file: sec/5_conclusion.tex
\section{Conclusion}
\label{sec:conclusion}

We propose StairVLA, a stage-aware hierarchical action generation framework that combines low-frequency high-level generation with high-frequency observation-conditioned refinement to reduce inference cost while maintaining strong task performance. More importantly, our results suggest that VLA action generation is not a homogeneous denoising process: different stages exhibit different information dependencies, while partially denoised actions themselves already encode task-relevant information for subsequent refinement. Our method exploits this stage difference by explicitly separating early-stage generation from later-stage refinement. Further understanding the functional roles of different denoising stages, better utilizing the information encoded in intermediate actions, and learning such hierarchical structures adaptively remain important directions for future research.

%% file: sec/6_statement.tex
\section*{Funding and Disclaimer}

This work received no funding from Ant Technology U.S., Inc. or Ant Group.
The views expressed in this paper are those of the authors and do not necessarily
reflect the views of Ant Technology U.S., Inc. or Ant Group.

%% file: sec/a1_details.tex
\section{Implementation Details}
\label{section:a1}

Unless otherwise specified, all experiments use the model architecture, training protocol, and evaluation settings described in this section.

\subsection{Model Architecture}

\paragraph{High-level policy.}
The high-level policy uses Qwen3-VL-4B-Instruct as the vision-language backbone, followed by a GR00T-style flow-matching action head. The action prediction horizon is set to \(H=32\).

\paragraph{Low-level refiner.}
The low-level refiner adopts a DiT-style flow-matching architecture with 16 Transformer blocks, hidden dimension 768, 12 attention heads, and FFN dimension 3072. Cross-attention and self-attention blocks are interleaved. The refiner predicts an action chunk of length \(h=5\).

\paragraph{Visual and high-level conditioning.}
The refiner uses a trainable SigLIP-B/16 visual encoder initialized from \texttt{google/siglip-base-patch16-224}, with RGB observations from both the primary and wrist cameras. The final-layer hidden states of the high-level VLM are first projected to dimension 768 and then compressed along the token dimension using adaptive average pooling, producing eight high-level latent tokens. These features are cached together with the high-level action trajectory and reused across multiple low-level refinement cycles.

\paragraph{Local action context.}
For each query chunk, the refiner additionally uses five coarse actions before and five coarse actions after the query segment from the same high-level trajectory as local temporal context. The query chunk itself is excluded from the context. A learned padding token is used when the required context extends beyond the trajectory boundary.

\paragraph{Multi-rate inference.}
The refiner is invoked once every \(h=5\) environment steps and returns five actions that are executed sequentially. The high-level policy is refreshed every \(M\) low-level refinement cycles, corresponding to \(5M\) environment steps. Refined actions are not written back into the cached high-level trajectory; each local chunk is refined from the original high-level coarse trajectory.

\subsection{Training Details}

\paragraph{High-level policy training.}
The high-level policy is trained on the combined LIBERO dataset for 30k steps with an action horizon of \(H=32\). After training, both the high-level VLM and action head are frozen during refiner training.

\paragraph{Refiner training.}
During refiner training, the frozen high-level policy remains active to generate intermediate action trajectories. The high-level policy is run to a fixed denoising progress \(\alpha=0.8\), producing a partially denoised trajectory \(C\).

For each training sample, we then sample
\[
\tau \sim \mathcal{U}(0.8,1.0),
\]
and construct the refiner input by interpolating between the high-level coarse trajectory and the ground-truth action:
\[
x_\tau
=
C
+
\frac{\tau-\alpha}{1-\alpha}
\left(
a_{\mathrm{gt}}-C
\right).
\]
The refiner predicts the remaining flow velocity toward the ground-truth action and is trained using a validity-masked mean-squared error. The refiner is trained for 30k steps.

\paragraph{Optimization.}
The refiner is trained with AdamW using a learning rate of \(5\times10^{-5}\) for the refiner head and a weight decay of \(0.01\). We use a cosine learning-rate schedule with 1k warmup steps. Training uses bf16 mixed precision, DeepSpeed ZeRO-2, and gradient clipping, and is conducted on eight NVIDIA RTX A6000 48GB GPUs. The per-GPU batch size is 16 with no gradient accumulation, resulting in a global batch size of 128.

\paragraph{Dataset and preprocessing.}
We train on the combined LIBERO dataset and evaluate on LIBERO-Spatial, LIBERO-Object, LIBERO-Goal, and LIBERO-Long. The action dimension is 7 and the high-level action horizon is 32. Continuous action dimensions are min--max normalized to \([-1,1]\). Action sequences extending beyond episode boundaries are zero-padded, and invalid padded steps are excluded from the training loss using a validity mask. Both primary and wrist-camera observations are used during training and evaluation, and video frames are decoded using \texttt{torchvision\_av}.

%% file: sec/a2_experiments_all.tex
\section{Additional Experimental Results and Analysis}
\label{section:a2}

\subsection{Denoising-Stage Conditioning Analysis}
\label{sec:denoising_stage_conditioning}

We further analyze how generative VLA action heads use different conditioning information throughout the denoising process. Specifically, we examine how action queries allocate cross-attention to different conditioning sources across denoising stages in both GR00T and $\pi_0$. This analysis is intended as a diagnostic of stage-dependent conditioning behavior rather than a strict causal attribution of individual token groups to the final action prediction.

\subsubsection{Experimental Setup}

Here, we use two trained StarVLA checkpoints based on GR00T and $\pi_0$, and randomly sample 16 observation--instruction pairs from the LIBERO LeRobot dataset. All analyses are performed at inference time without additional training or parameter updates.

To examine how attention evolves throughout denoising, we densify the inference trajectory into 100 analysis stages and record the cross-attention probability from action queries to conditioning-token keys at each stage. This 100-stage trajectory is used only for analysis and visualization and does not correspond to the native inference setting of the models, whose default inference trajectory uses four denoising steps. For each model, we analyze both the first and last action-head layers, resulting in four cases: GR00T first layer, GR00T last layer, $\pi_0$ first layer, and $\pi_0$ last layer.

At each denoising stage, we first average the attention probabilities over attention heads and action-query tokens to obtain a key-side attention distribution. We then sum the attention mass over tokens belonging to the same group and normalize it by the total attention mass over valid tokens. The reported curves represent the mean attention mass over the 16 sampled observation--instruction pairs.

We divide the conditioning tokens into five groups: \emph{task instruction}, \emph{wrapper prompt}, \emph{grounding prompt}, \emph{image tokens}, and \emph{special tokens}. The task instruction corresponds to the actual LIBERO command, such as ``pick up the cream cheese and place it in the basket.'' The wrapper prompt refers to template text preceding the task instruction, such as ``Your task is.'' The grounding prompt refers to localization-oriented template text following the task instruction, such as a request to identify key objects using bounding-box coordinates. Importantly, the grounding prompt specifies only the output format and does not contain ground-truth bounding boxes. Image tokens correspond to visual inputs, while special tokens include chat-template tokens, image-boundary tokens, role-boundary tokens, padding tokens, and other special tokens.

For compact visualization, wrapper-prompt, grounding-prompt, and special tokens can be grouped as ``other prompt tokens.'' In the analysis below, we keep the five groups separate to better examine the source of attention shifts across models.

\subsubsection{Cross-Attention Allocation across Denoising Stages}

\begin{figure*}[t]
    \centering
    \includegraphics[width=\textwidth]{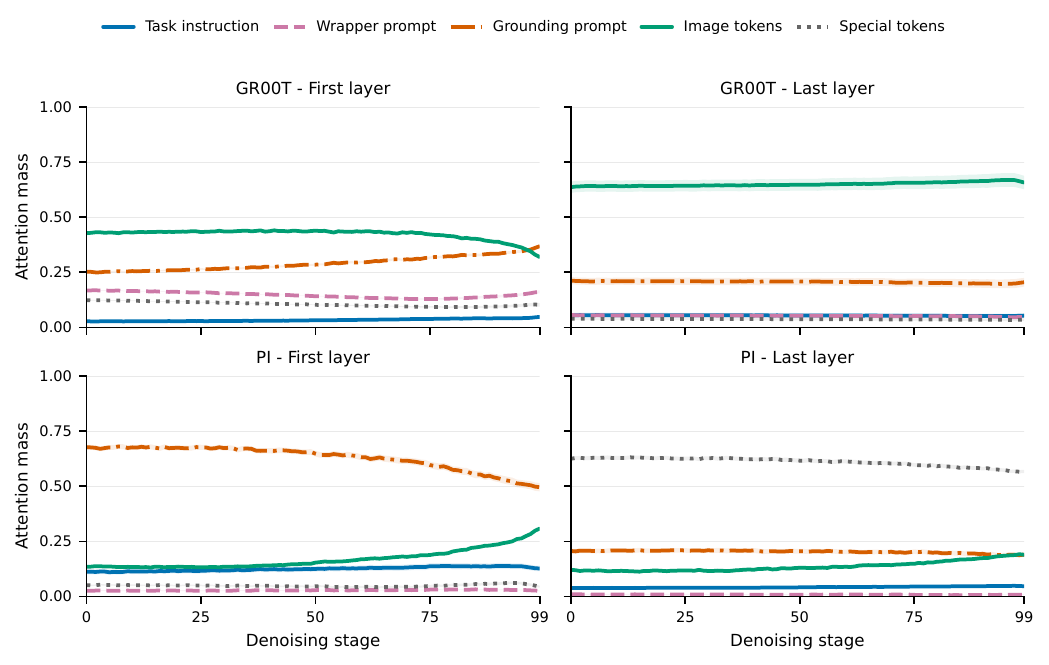}
    \caption{
    Denoising-stage attention analysis for GR00T and $\pi_0$.
    We record action-query cross-attention to different conditioning-token groups over 100 dense denoising stages.
    Both models exhibit stage-dependent attention patterns, with the changes being more pronounced in the first action-head layer.
    }
    \label{fig:appendix_attention_analysis}
    \vspace{-0.3cm}
\end{figure*}

The overall results are shown in Fig.~\ref{fig:appendix_attention_analysis}. Both GR00T and $\pi_0$ exhibit changes in cross-attention allocation throughout denoising, indicating that the action head does not read conditioning information in a completely static manner during generation. Compared with the last layer, these changes are more pronounced in the first action-head layer.

The first-layer result of $\pi_0$ provides the clearest example. As denoising approaches later stages, the attention mass allocated to image tokens increases substantially, while the proportion assigned to prompt-related tokens decreases. This suggests that late-stage action queries in $\pi_0$ place greater explicit emphasis on visual conditioning, consistent with the interpretation that later refinement increasingly focuses on alignment with the current observation.

GR00T exhibits a different surface trend. In its first action-head layer, image-token attention decreases toward later stages, while attention to grounding-prompt and other prompt-related positions increases. We attribute this difference primarily to the different conditioning architectures of the two models. $\pi_0$ uses layer-wise VLM conditioning, such that its first action-head layer receives comparatively shallow VLM-contextualized representations. In contrast, GR00T conditions its action head on the final-layer hidden states of Qwen2.5-VL. Therefore, even at the first action-head layer, token positions corresponding to grounding or prompt text have already undergone multiple layers of visual-language contextualization.

Consequently, the increase in attention to grounding-related token positions in GR00T should not be interpreted simply as an increased reliance on template text. Importantly, these grounding representations themselves encode image-derived spatial information, such as object bounding boxes. Their increased attention therefore more likely reflects a reallocation of cross-attention toward more explicit visual-spatial grounding representations during later denoising stages. Overall, although GR00T and $\pi_0$ exhibit different attention patterns, both suggest a shift toward more visually grounded information in later denoising stages.

The last-layer trends are weaker and should be interpreted more cautiously. One possible explanation is that the first action-head layer is the earliest point at which action queries explicitly read conditioning tokens, making it more sensitive to stage-dependent changes in conditioning usage. After multiple action-head transformations, conditioning information has progressively been integrated into the evolving action representation, making explicit attention over the original token groups less sensitive to such changes. The last layer of $\pi_0$ still exhibits a weaker but visible increase in image-token attention, whereas GR00T remains comparatively stable.

\subsubsection{Qualitative Attention Maps and Self-Attention}

\begin{figure*}[t]
    \centering

    \begin{minipage}{0.98\textwidth}
        \centering
        \includegraphics[width=\textwidth]{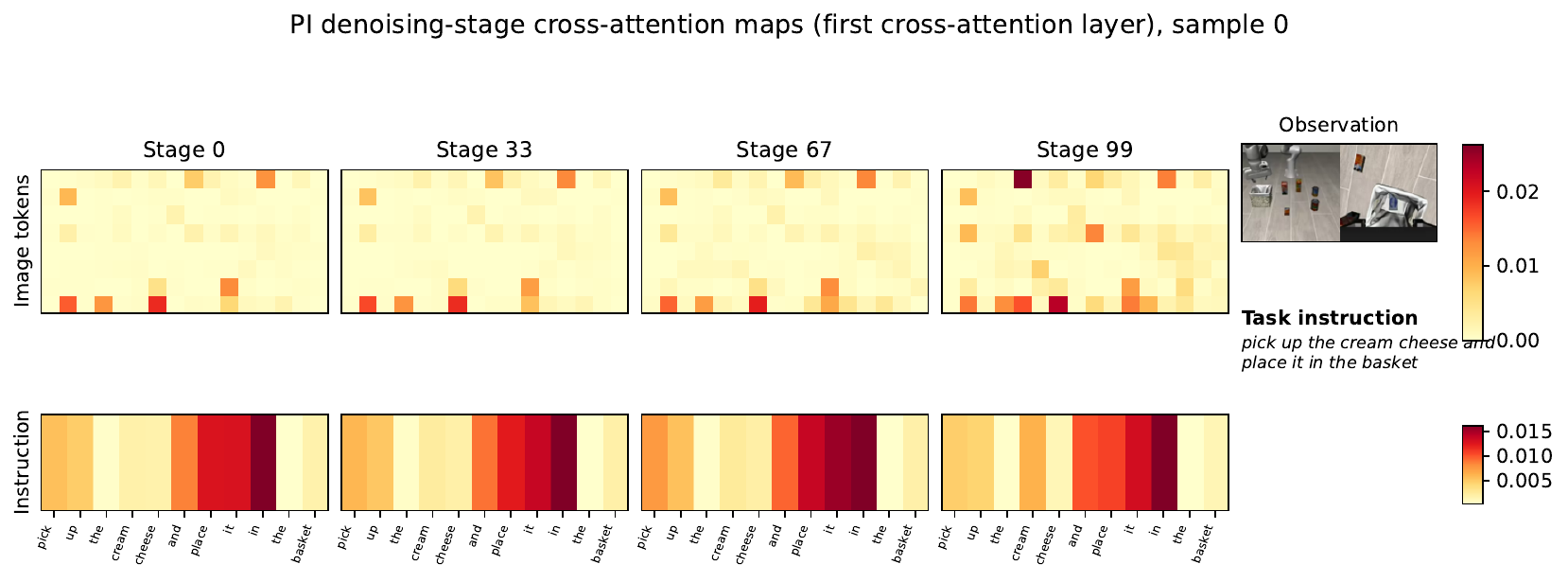}
        \vspace{-0.5em}
        \centerline{(a) $\pi_0$}
    \end{minipage}

    \vspace{0.8em}

    \begin{minipage}{0.98\textwidth}
        \centering
        \includegraphics[width=\textwidth]{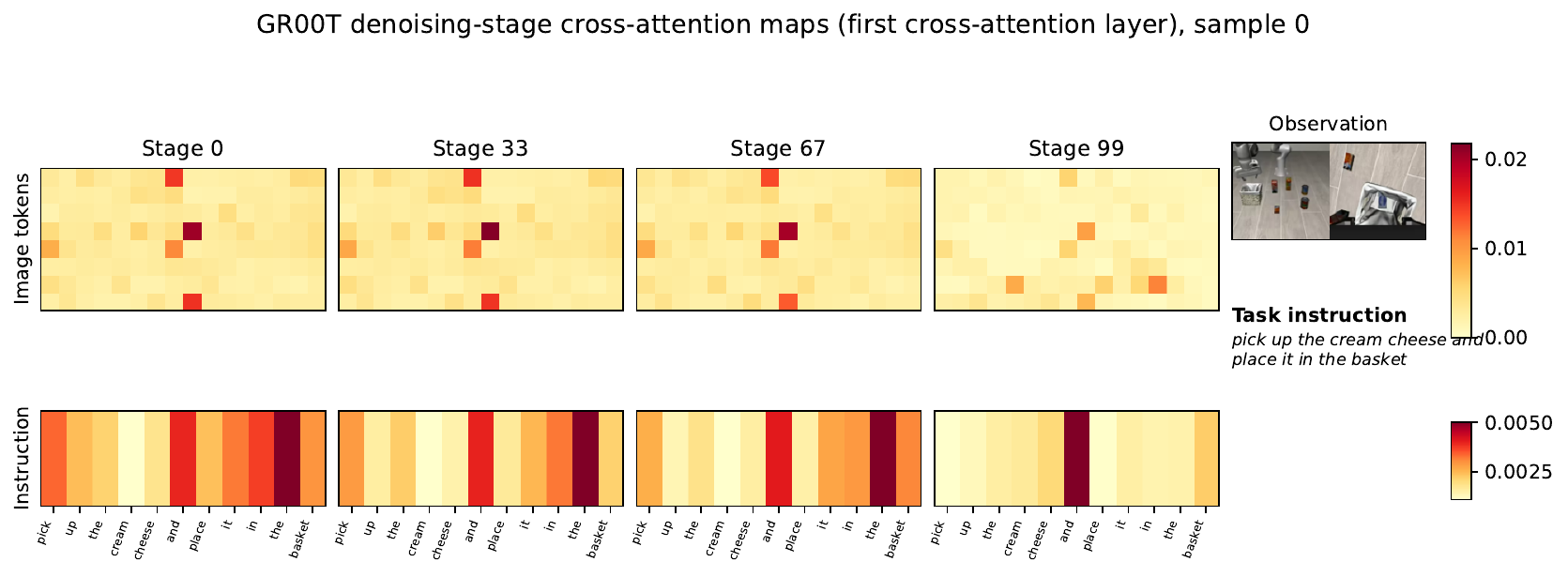}
        \vspace{-0.5em}
        \centerline{(b) GR00T}
    \end{minipage}

    \caption{
    Representative first-layer cross-attention maps of $\pi_0$ and GR00T at different denoising stages.
    For $\pi_0$, attention to image tokens becomes stronger toward later stages, while task-instruction attention does not exhibit a clear monotonic trend.
    For GR00T, late-stage cross-attention does not simply shift toward raw image or task-instruction tokens, indicating that different VLA action heads may realize stage-dependent conditioning through different token pathways.
    }
    \label{fig:pi_gr00t_first_layer_attention_maps}
\end{figure*}

\begin{figure}[t]
    \centering
    \includegraphics[width=0.62\linewidth]{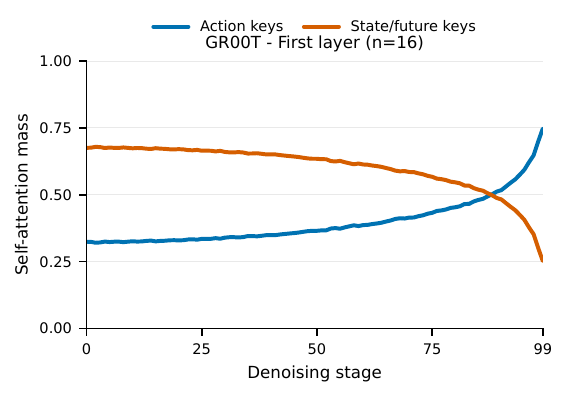}
    \caption{
    First-layer self-attention allocation in GR00T across denoising stages.
    Attention to action keys increases toward later stages, while attention to state/future keys decreases.
    }
    \label{fig:gr00t_self_attention}
\end{figure}

To complement the averaged attention curves, we further visualize representative first-layer attention maps. As shown in Fig.~\ref{fig:pi_gr00t_first_layer_attention_maps}(a), $\pi_0$ assigns stronger attention to image tokens at later denoising stages, particularly near the end of the trajectory, while task-instruction attention does not show a clear monotonic trend. This qualitative observation is consistent with the averaged curves and further illustrates the increased explicit emphasis on visual information during late-stage denoising.

In contrast, Fig.~\ref{fig:pi_gr00t_first_layer_attention_maps}(b) shows that GR00T does not simply shift its late-stage cross-attention toward raw image or task-instruction tokens. In this representative example, both image-token and task-instruction attention become weaker near the final stage. Together with the averaged results in Fig.~\ref{fig:appendix_attention_analysis}, this suggests that GR00T reallocates late-stage attention toward other contextualized token positions, such as grounding- or template-related representations. These results further indicate that different VLA action heads can exhibit stage-dependent conditioning behavior through different token pathways.

In addition to cross-attention, we analyze self-attention within the GR00T action head. As shown in Fig.~\ref{fig:gr00t_self_attention}, the attention mass assigned to action keys increases throughout denoising, while the attention to state/future keys decreases. This suggests that later denoising increasingly operates on the evolving action representation itself. One possible interpretation is that, once task-level action structure has formed, subsequent denoising increasingly refines the existing action representation, while external conditioning information may already have been selectively read and integrated at earlier stages or shallower action-head layers.

Finally, we emphasize that attention mass does not provide a strict causal attribution of conditioning information. It measures only the explicit attention allocated to different key-token groups at a particular layer and denoising stage, rather than their causal contribution to the final action prediction. Stronger causal evidence requires intervention-based analyses, such as masking image or instruction tokens and measuring the resulting change in action prediction. Nevertheless, the attention analysis consistently shows that both GR00T and $\pi_0$ exhibit denoising-stage-dependent conditioning patterns, providing additional empirical evidence for stage-dependent conditioning behavior in generative VLA action formation.

\subsection{Generalization Across VLA Backbones}
\label{sec:backbone_transfer}

The main experiments report the overall results of our method on both GR00T-style and $\pi$-style generative VLA backbones. Here, we provide the complete LIBERO results under different high-level reuse horizons, including per-suite success rates and native inference latency.

As shown in Table~\ref{tab:backbone_transfer}, our method maintains strong task performance across different reuse horizons on both backbones while substantially reducing inference cost. For the GR00T-style backbone, the \(5{\times}4\) setting achieves the highest average success rate of \(97.8\%\), while the \(5{\times}5\) setting still maintains \(97.05\%\) success with lower latency. For the $\pi$-style backbone, performance improves from \(97.8\%\) for the StarVLA-$\pi$ baseline to \(98.3\%\) at \(5{\times}4\), with a native latency of \(44.5\) ms per 5-action chunk.

These results further support the cross-backbone generality of the proposed hierarchical refinement framework.

\begin{table*}[!t]
\centering
\small
\setlength{\tabcolsep}{4.0pt}
\renewcommand{\arraystretch}{1.12}
\caption{
Complete LIBERO results across GR00T-style and $\pi$-style VLA backbones.
\(M\) denotes the number of 5-action low-level chunks executed between two high-level policy updates.
The hierarchical GR00T- and $\pi$-based models use high-level action horizons of \(H=32\) and \(H=20\), respectively.
Latency is reported per native returned action chunk.
The non-hierarchical StarVLA baselines use 8-action chunks, whereas our low-level refiner outputs 5-action chunks.
}
\label{tab:backbone_transfer}
\resizebox{\textwidth}{!}{
\begin{tabular}{l c c cccc c c c}
\toprule
\textbf{Method}
& \textbf{\(M\)}
& \textbf{Chunk Size}
& \textbf{Spatial}
& \textbf{Object}
& \textbf{Goal}
& \textbf{Long}
& \textbf{Avg.}
& \textbf{\(\Delta\) Avg.}
& \textbf{Latency} \\
\midrule

\multicolumn{10}{c}{\textit{GR00T-style backbone}} \\
\midrule

StarVLA-GR00T
& --
& 8
& 97.8 & 98.8 & 97.4 & 92.0
& 96.5 & -- & 115.0 \\

Ours (GR00T-base)
& 2
& 5
& 98.4 & 98.8 & 97.6 & 94.8
& 97.40 & +0.90 & 68.7 \\

Ours (GR00T-base)
& 3
& 5
& \textbf{99.2} & 98.8 & \textbf{97.8} & 94.0
& 97.45 & +0.95 & 52.3 \\

\rowcolor{gray!15}
Ours (GR00T-base)
& 4
& 5
& 98.0 & \textbf{99.6} & \textbf{97.8} & \textbf{95.8}
& \textbf{97.80} & \textbf{+1.30} & 44.2 \\

Ours (GR00T-base)
& 5
& 5
& 98.4 & 98.8 & 97.4 & 93.6
& 97.05 & +0.55 & \textbf{39.3} \\

\midrule
\multicolumn{10}{c}{\textit{$\pi$-style backbone}} \\
\midrule

StarVLA-$\pi$
& --
& 8
& 99.2 & 99.0 & \textbf{97.2} & 95.8
& 97.8 & -- & 183.3 \\

Ours ($\pi$-base)
& 2
& 5
& 99.2 & 99.4 & 96.2 & 95.8
& 97.7 & -0.1 & 70.7 \\

Ours ($\pi$-base)
& 3
& 5
& 99.4 & \textbf{99.6} & 96.8 & 96.2
& 98.0 & +0.2 & 53.1 \\

\rowcolor{gray!15}
Ours ($\pi$-base)
& 4
& 5
& \textbf{99.6} & 99.0 & 96.8 & \textbf{97.4}
& \textbf{98.3} & \textbf{+0.5} & \textbf{44.5} \\

\bottomrule
\end{tabular}
}
\end{table*}

\subsection{Long-Horizon Action Generation and Reuse}
\label{sec:long_horizon_reuse}

We further investigate how the proposed hierarchical architecture behaves when the high-level policy is extended to substantially longer action horizons.
In addition to the default setting, we train high-level GR00T policies with action horizons of \(H=48\) and \(H=64\), together with their corresponding low-level refiners.
The refiner predicts 5-step action chunks, and the cached high-level trajectory is refreshed after executing \(M\) consecutive refinement chunks.
Therefore, one high-level policy invocation corresponds to an effective execution horizon of \(5M\) actions.
For \(H=48\) and \(H=64\), the maximum refinement horizons used during training are \(5{\times}9=45\) and \(5{\times}12=60\) actions, respectively.

\begin{figure}[t]
    \centering
    \includegraphics[width=0.35\linewidth]{
        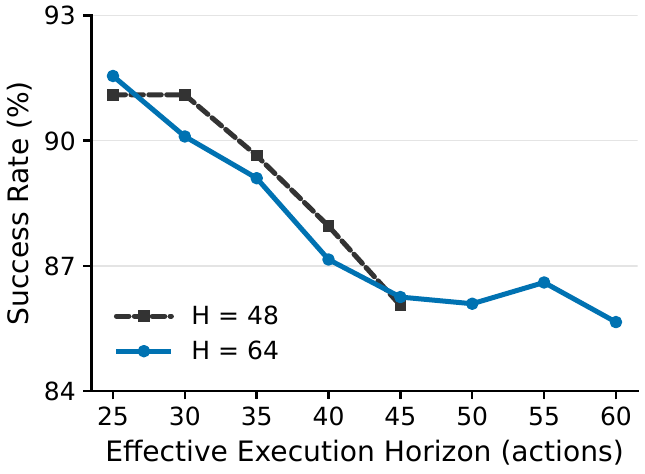
    }
    \caption{
    Effect of long-horizon high-level trajectory reuse.
    We report the average LIBERO success rate as the cached high-level trajectory is reused for increasingly long execution horizons.
    Both curves are shown from 25 executed actions onward, corresponding to the shared evaluation range of the \(H=48\) and \(H=64\) models.
    }
    \label{fig:long_horizon_reuse}
    \vspace{-0.5cm}
\end{figure}

As shown in Fig.~\ref{fig:long_horizon_reuse}, task performance generally decreases as the same high-level trajectory is reused for a longer period.
For \(H=48\), the average success rate remains at \(91.10\%\) for execution horizons of 25 and 30 actions, but gradually decreases to \(86.05\%\) when the trajectory is reused for 45 actions.
A similar trend is observed for \(H=64\): the success rate decreases from \(91.55\%\) at 25 actions and remains around \(85\%\)--\(87\%\) when the reuse horizon is extended to 40--60 actions.
Notably, increasing the high-level prediction horizon from \(H=48\) to \(H=64\) does not provide a clear performance advantage under long reuse.
This suggests that simply training the upper-level policy to predict farther into the future does not directly translate into a proportionally longer effective reuse horizon.

These results reveal an important limitation of the current hierarchical action-generation architecture: although a high-level trajectory can be reused across multiple low-level refinement cycles, its effective reuse range is not unlimited.
When the high-level policy generates and caches an increasingly long future plan, the subsequent execution must continue to operate under an evolving closed-loop state.
As execution proceeds, uncertainty arising from action errors, observation changes, and environment interactions may accumulate, causing the cached high-level trajectory to become progressively less consistent with the current state.
Although the low-level refiner can use the latest visual observation to locally adjust the executable actions, its refinement remains conditioned on the cached high-level trajectory.
Consequently, when the upper-level plan spans an excessively long timescale and is reused without refresh, local refinement may no longer be sufficient to compensate for the growing mismatch between the cached plan and the actual closed-loop state.

From this perspective, the efficiency benefit of hierarchical VLA inference cannot be obtained simply by continuously increasing the high-level prediction horizon or reuse length.
A longer cached trajectory further amortizes the expensive high-level VLA invocation, but simultaneously reduces the frequency at which the high-level decision process can respond to newly observed states.
Our results therefore suggest a practical trade-off between high-level reuse and closed-loop responsiveness:
within a moderate reuse range, low-level refinement can effectively provide local closed-loop correction while reducing high-level inference frequency;
beyond this range, accumulated uncertainty and plan--state mismatch may exceed what the local refiner can compensate for, leading to substantial performance degradation.

This observation also motivates a natural direction for future work.
Rather than refreshing the high-level trajectory after a fixed number of refinement cycles, an adaptive mechanism could trigger a new high-level VLA invocation according to the observed state deviation, action uncertainty, or consistency between the cached trajectory and the current observation.
Such an event-driven or uncertainty-aware high-level refresh strategy may provide a more flexible balance between inference efficiency and closed-loop robustness.

%% file: sec/a3_realworld.tex
\section{Real-World Experimental Details}
\label{section:a3}

\subsection{Real-World Platform and System Setup}

We conduct real-world experiments on an AgileX PiperX 6-DoF robotic arm. The system uses a fixed top camera and a wrist-mounted Intel RealSense camera to provide dual-view RGB observations at a resolution of \(640\times480\). The policy takes as input the language instruction, dual-view images, and an 8-dimensional robot state
\[
s_t = [x, y, z, q_x, q_y, q_z, q_w, g],
\]
where \((x,y,z)\) denotes the end-effector position, \((q_x,q_y,q_z,q_w)\) represents the end-effector orientation as a quaternion, and \(g\) denotes the gripper opening.

The policy outputs a 7-dimensional end-effector action
\[
a_t = [\Delta x, \Delta y, \Delta z, \Delta r_x, \Delta r_y, \Delta r_z, g^*],
\]
where the translational delta is defined in the robot world frame, the rotational delta is represented as an axis-angle increment in the local end-effector frame, and \(g^*\) denotes the absolute gripper target. During execution, the predicted end-effector target pose is converted to joint targets through local inverse kinematics.

For safe and consistent real-world evaluation, all methods use the same control constraints: the per-step translation and rotation are limited to \(0.05\,\mathrm{m}\) and \(0.20\,\mathrm{rad}\), respectively, and the maximum joint velocity is limited to \(25^\circ/\mathrm{s}\). Before each trial, the robot returns to a fixed initial configuration, the internal action cache is cleared, and the cameras are warmed up.

Fruit25 and PushBlock are executed at control frequencies of 8~Hz and 20~Hz, respectively. Inference is synchronous: when the action buffer is empty, the control loop waits for a new action chunk and then executes the predicted actions sequentially at the corresponding control frequency. Independently of the policy control frequency, robot feedback is recorded at approximately 50~Hz for subsequent trajectory-quality and motion-smoothness analysis. All real-world experiments are conducted on the same NVIDIA RTX A4000 GPU.

\subsection{Datasets and Data Processing}

\begin{figure*}[!t]
\centering

\begin{minipage}[t]{0.8\textwidth}
    \centering
    \includegraphics[width=\linewidth]{
        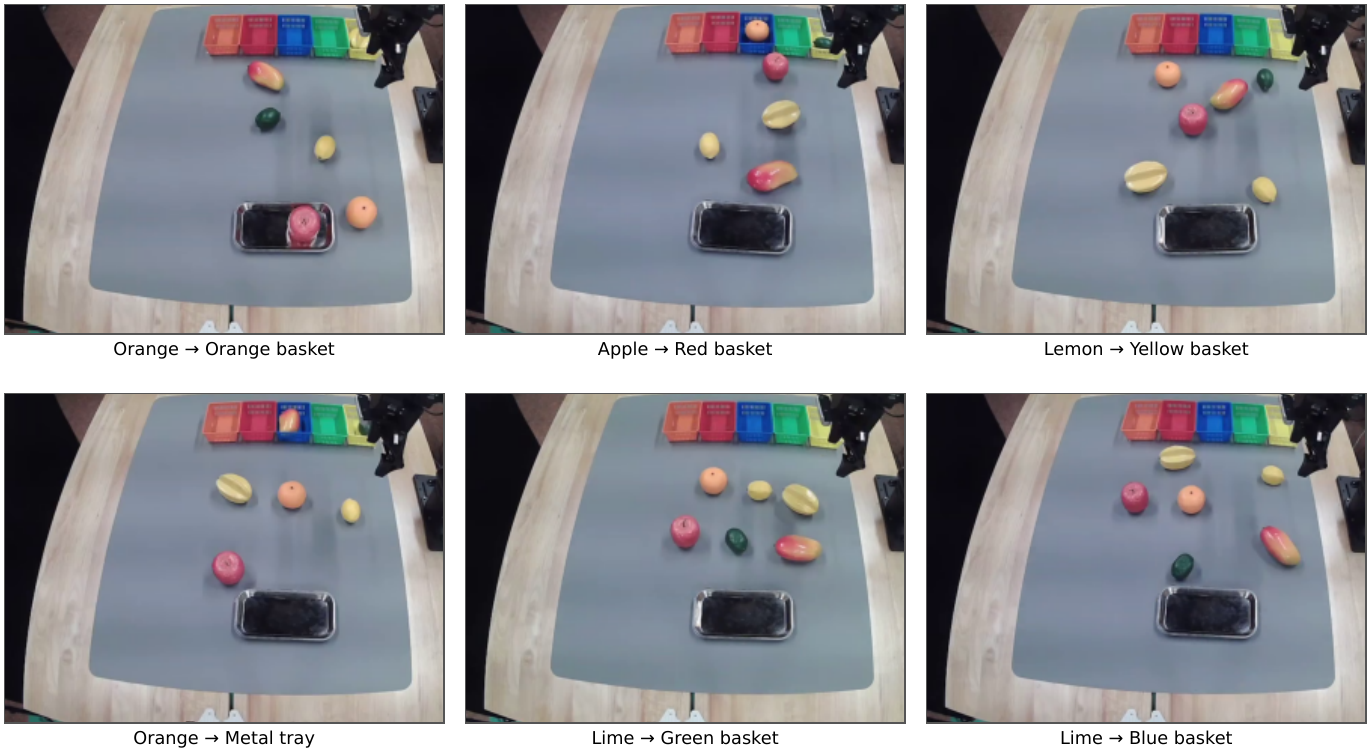
    }
    \captionof{figure}{
    Example initial scenes from Fruit25.
    }
    \label{fig:fruit25_scene_examples}
\end{minipage}
\hfill
\begin{minipage}[t]{0.8\textwidth}
    \centering
    \includegraphics[width=\linewidth]{
        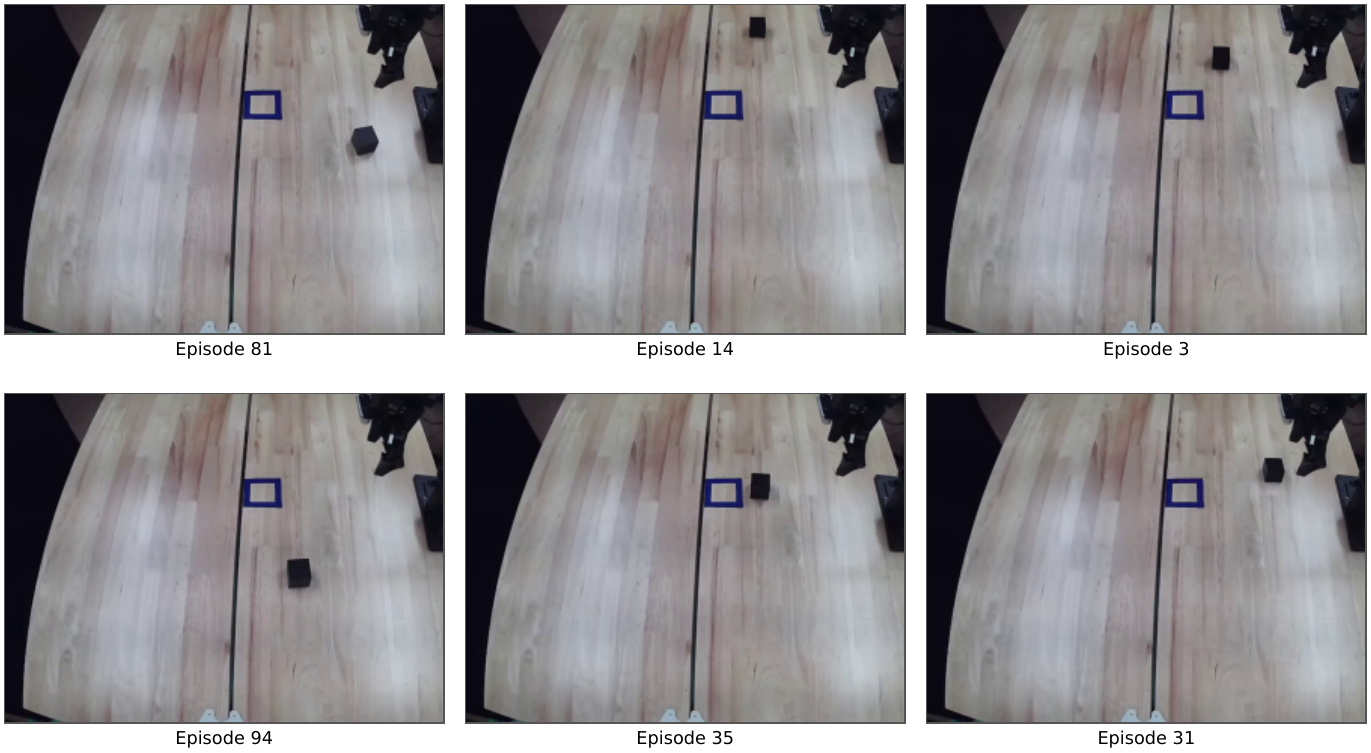
    }
    \captionof{figure}{
    Example initial scenes from PushBlock.
    }
    \label{fig:pushblock_scene_examples}
\end{minipage}

\end{figure*}

The two real-world environments are illustrated in
Figs.~\ref{fig:fruit25_scene_examples} and~\ref{fig:pushblock_scene_examples}. Fruit25 is collected through VR teleoperation and contains 25 tasks. Among them, 24 are single-fruit pick-and-place tasks formed by four fruit categories---apple, orange, lemon, and lime---and six target locations, including five colored baskets and one metal tray. The final task is a long-horizon multi-object fruit-sorting task. We perform quality control on 1,220 collected episodes through trajectory inspection, video inspection, and manual replay review, removing episodes with clear execution errors, corrupted recordings, or abnormal trajectory segments. This results in 1,186 valid demonstrations with 109,684 frames. PushBlock contains a single continuous pushing task in which the robot pushes a black block into a blue target region. The dataset contains 100 demonstrations with 32,776 frames. Both Fruit25 and PushBlock include top and wrist RGB observations, end-effector poses, gripper states, and language instructions.

The two datasets use different sampling and control frequencies. Fruit25 mainly consists of relatively discrete and quasi-static manipulation behaviors such as grasping, transporting, and placing objects, for which 8~Hz provides sufficient temporal resolution. In contrast, PushBlock requires sustained contact with the object and more frequent closed-loop corrections during execution, so we use a higher frequency of 20~Hz to provide finer-grained motion control.

All demonstrations are recorded only after robot initialization is completed, so initialization motion is excluded from the training trajectories. We do not apply a unified automatic trimming procedure to remove stationary segments at the beginning or end of trajectories; instead, low-quality episodes are removed as complete trajectories. The recorded end-effector poses are converted into temporal delta-EE actions. The translational increment is defined as
\[
\Delta p_t = p_{t+1} - p_t,
\]
and the rotational increment is defined as
\[
\Delta r_t = \log\left(R_t^{-1}R_{t+1}\right).
\]
The gripper command uses the absolute opening at the next timestep, and the motion delta of the final frame is set to zero.

\begin{table*}[t]
\centering
\small
\setlength{\tabcolsep}{4.5pt}
\renewcommand{\arraystretch}{1.1}
\caption{
Task definitions and retained demonstration counts for the real-world
datasets.
}
\label{tab:realworld_tasks}
\resizebox{\textwidth}{!}{
\begin{tabular}{c c l c}
\toprule
\textbf{Task ID} & \textbf{Benchmark} & \textbf{Task Description} & \textbf{Retained Demonstrations} \\
\midrule
1  & Fruit25 & Pick up the apple and place it into the red basket. & 100 \\
2  & Fruit25 & Pick up the apple and place it onto the metal tray. & 100 \\
3  & Fruit25 & Pick up the orange and place it onto the metal tray. & 100 \\
4  & Fruit25 & Pick up the orange and place it into the orange basket. & 96 \\
5  & Fruit25 & Pick up the lemon and place it onto the metal tray. & 98 \\
6  & Fruit25 & Pick up the lemon and place it into the yellow basket. & 99 \\
7  & Fruit25 & Pick up the lime and place it into the green basket. & 98 \\
8  & Fruit25 & Pick up the apple and place it into the orange basket. & 20 \\
9  & Fruit25 & Pick up the apple and place it into the blue basket. & 20 \\
10 & Fruit25 & Pick up the apple and place it into the green basket. & 19 \\
11 & Fruit25 & Pick up the apple and place it into the yellow basket. & 20 \\
12 & Fruit25 & Pick up the orange and place it into the red basket. & 19 \\
13 & Fruit25 & Pick up the orange and place it into the blue basket. & 20 \\
14 & Fruit25 & Pick up the orange and place it into the green basket. & 20 \\
15 & Fruit25 & Pick up the orange and place it into the yellow basket. & 20 \\
16 & Fruit25 & Pick up the lemon and place it into the orange basket. & 20 \\
17 & Fruit25 & Pick up the lemon and place it into the red basket. & 20 \\
18 & Fruit25 & Pick up the lemon and place it into the blue basket. & 20 \\
19 & Fruit25 & Pick up the lemon and place it into the green basket. & 20 \\
20 & Fruit25 & Pick up the lime and place it into the orange basket. & 20 \\
21 & Fruit25 & Pick up the lime and place it into the red basket. & 20 \\
22 & Fruit25 & Pick up the lime and place it into the blue basket. & 20 \\
23 & Fruit25 & Pick up the lime and place it into the yellow basket. & 13 \\
24 & Fruit25 & Pick up the lime and place it onto the metal tray. & 84 \\
25 & Fruit25 & Sort the fruits into their corresponding baskets. & 100 \\
\midrule
1 & PushBlock & Push the black block into the blue target region. & 100 \\
\bottomrule
\end{tabular}
}
\end{table*}

\subsection{Model Configuration and Training}
\label{sec:realworld_model_training}

We compare StarVLA-GR00T, StarVLA-$\pi$, and our hierarchical model. All three methods use Qwen3-VL-2B-Instruct as the vision-language backbone and are trained on the same data, language instructions, and state/action representations for each benchmark. StarVLA-GR00T and StarVLA-$\pi$ are fully fine-tuned end-to-end and use an action horizon of 8.

For our method, we set the high-level action horizon to \(H=32\), the low-level chunk size to \(h=8\), and the reuse length to \(M=3\), corresponding to 24 executed actions between two high-level policy updates. During refiner training, the high-level partial-denoising progress is sampled from \([0.6,1.0]\), the low-level refinement scale from \([0.8,1.0]\), and the temporal-context trajectory is fully denoised. At inference time, we use a high-level denoising progress of \(0.97\), two high-level inference steps, and one low-level refinement step, with the refiner operating from an assumed denoising progress of \(0.97\).

For Fruit25, all models are trained for 30k steps, while for PushBlock they are trained for 10k steps. On Fruit25, StarVLA-GR00T and both stages of our method use a global batch size of 64, while StarVLA-$\pi$ uses a global batch size of 32. On PushBlock, all three methods use an effective batch size of 8. We use a cosine learning-rate schedule with 1,000 warm-up steps, a main learning rate of \(10^{-4}\) for the action-related modules, and \(2.5\times10^{-5}\) for the vision-language backbone.

\subsection{Evaluation Protocol}
\label{sec:realworld_evaluation}

All methods are evaluated in the same real-world environment with fixed camera poses, lighting conditions, background, and robot initial configuration. Since object placement is performed manually, identical object poses cannot be guaranteed across methods; instead, we maintain similar initial layouts across trials.

For Fruit25, Tasks 1--7, 24, and 25 are evaluated with 10 trials per method, while Tasks 8--23 are evaluated with 5 trials per method, resulting in 170 trials for each method. For PushBlock, each method is evaluated with 10 trials, which may be terminated earlier once the task is clearly completed or has failed.

All tasks use manually assigned binary success/failure labels. For the atomic Fruit25 tasks, a trial is considered successful if the specified fruit is grasped and placed into the instructed basket or onto the metal tray. For the Sort Fruits task, four fruits are present in the scene, and the trial is considered successful if at least three are correctly sorted into their corresponding destinations. For PushBlock, success is defined as pushing the black block into the blue target region.

\begin{table*}[htbp]
\centering
\small
\setlength{\tabcolsep}{3.8pt}
\renewcommand{\arraystretch}{1.05}
\caption{
Per-task success results on the real-world benchmarks.
Fruit25 tasks are evaluated with either 5 or 10 trials, while PushBlock is evaluated with 10 trials per method.
}
\label{tab:realworld_per_task_results}
\resizebox{\textwidth}{!}{
\begin{tabular}{c l c c c}
\toprule
\textbf{ID}
& \textbf{Task}
& \textbf{StarVLA-GR00T}
& \textbf{StarVLA-$\pi$}
& \textbf{Ours} \\
\midrule

\multicolumn{5}{c}{\textit{Fruit25}} \\
\midrule

1  & Apple $\rightarrow$ Red basket
& 7/10 (70\%) & 6/10 (60\%) & \textbf{10/10 (100\%)} \\

2  & Apple $\rightarrow$ Metal tray
& \textbf{10/10 (100\%)} & 9/10 (90\%) & 9/10 (90\%) \\

3  & Orange $\rightarrow$ Metal tray
& 9/10 (90\%) & 9/10 (90\%) & 9/10 (90\%) \\

4  & Orange $\rightarrow$ Orange basket
& 5/10 (50\%) & \textbf{8/10 (80\%)} & 7/10 (70\%) \\

5  & Lemon $\rightarrow$ Metal tray
& 10/10 (100\%) & 10/10 (100\%) & 10/10 (100\%) \\

6  & Lemon $\rightarrow$ Yellow basket
& 8/10 (80\%) & 8/10 (80\%) & 8/10 (80\%) \\

7  & Lime $\rightarrow$ Green basket
& 9/10 (90\%) & 9/10 (90\%) & \textbf{10/10 (100\%)} \\

8  & Apple $\rightarrow$ Orange basket
& \textbf{4/5 (80\%)} & 3/5 (60\%) & 3/5 (60\%) \\

9  & Apple $\rightarrow$ Blue basket
& 4/5 (80\%) & 4/5 (80\%) & \textbf{5/5 (100\%)} \\

10 & Apple $\rightarrow$ Green basket
& 3/5 (60\%) & 3/5 (60\%) & 3/5 (60\%) \\

11 & Apple $\rightarrow$ Yellow basket
& 2/5 (40\%) & \textbf{4/5 (80\%)} & 3/5 (60\%) \\

12 & Orange $\rightarrow$ Red basket
& 4/5 (80\%) & 4/5 (80\%) & \textbf{5/5 (100\%)} \\

13 & Orange $\rightarrow$ Blue basket
& 3/5 (60\%) & \textbf{5/5 (100\%)} & \textbf{5/5 (100\%)} \\

14 & Orange $\rightarrow$ Green basket
& \textbf{5/5 (100\%)} & 4/5 (80\%) & 3/5 (60\%) \\

15 & Orange $\rightarrow$ Yellow basket
& 3/5 (60\%) & \textbf{4/5 (80\%)} & 3/5 (60\%) \\

16 & Lemon $\rightarrow$ Orange basket
& \textbf{4/5 (80\%)} & 3/5 (60\%) & 1/5 (20\%) \\

17 & Lemon $\rightarrow$ Red basket
& 4/5 (80\%) & 4/5 (80\%) & \textbf{5/5 (100\%)} \\

18 & Lemon $\rightarrow$ Blue basket
& 4/5 (80\%) & \textbf{5/5 (100\%)} & \textbf{5/5 (100\%)} \\

19 & Lemon $\rightarrow$ Green basket
& 4/5 (80\%) & 4/5 (80\%) & \textbf{5/5 (100\%)} \\

20 & Lime $\rightarrow$ Orange basket
& \textbf{3/5 (60\%)} & \textbf{3/5 (60\%)} & 2/5 (40\%) \\

21 & Lime $\rightarrow$ Red basket
& \textbf{5/5 (100\%)} & 4/5 (80\%) & \textbf{5/5 (100\%)} \\

22 & Lime $\rightarrow$ Blue basket
& 5/5 (100\%) & 5/5 (100\%) & 5/5 (100\%) \\

23 & Lime $\rightarrow$ Yellow basket
& 3/5 (60\%) & 4/5 (80\%) & \textbf{5/5 (100\%)} \\

24 & Lime $\rightarrow$ Metal tray
& \textbf{10/10 (100\%)} & 6/10 (60\%) & \textbf{10/10 (100\%)} \\

25 & Sort Fruits
& \textbf{2/10 (20\%)} & 1/10 (10\%) & \textbf{2/10 (20\%)} \\

\midrule
\textbf{Overall}
& \textbf{170 trials}
& \textbf{130/170 (76.5\%)}
& \textbf{129/170 (75.9\%)}
& \textbf{138/170 (81.2\%)} \\

\midrule
\multicolumn{5}{c}{\textit{PushBlock}} \\
\midrule

1 & Push black block into blue target region
& 3/10 (30\%)
& 2/10 (20\%)
& \textbf{6/10 (60\%)} \\

\bottomrule
\end{tabular}
}
\end{table*}

\begin{figure}[htbp]
\centering
\begin{minipage}[t]{0.48\linewidth}
    \centering
    \includegraphics[width=\linewidth]{
        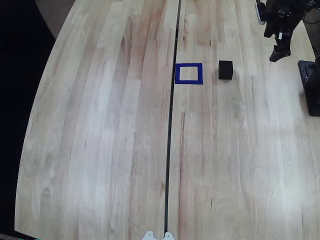
    }
\end{minipage}
\hfill
\begin{minipage}[t]{0.48\linewidth}
    \centering
    \includegraphics[width=\linewidth]{
        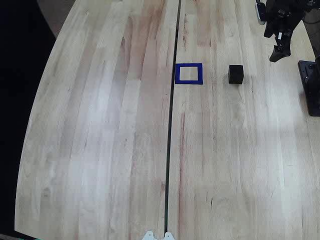
    }
\end{minipage}
\caption{
Examples of relatively easy PushBlock initial configurations, where the block is approximately aligned with the target region and can be pushed along a near-straight trajectory.
}
\label{fig:pushblock_easy_cases}
\end{figure}

\subsection{Detailed Main Results}
\label{sec:realworld_detailed_results}

We further report detailed success rates and latency results for the two benchmarks. Fruit25 is primarily composed of quasi-static pick-and-place tasks and is executed at 8~Hz; therefore, an 8-step action chunk corresponds to approximately 1~s of execution. As a result, inference is invoked relatively sparsely and the latency difference is less visually apparent during execution. In contrast, PushBlock runs at 20~Hz, where the same 8-step chunk covers only about 0.4~s, making inference latency more critical to continuous closed-loop control. In addition, real-world deployment introduces fixed overhead from image transmission, inverse kinematics, and robot control, which partially reduces the relative latency gain compared with simulation.

The advantage of our method is more pronounced on PushBlock. StarVLA-GR00T and StarVLA-$\pi$ mainly succeed in simple cases where the block is approximately aligned below the target region and can be pushed along a near-straight trajectory, as illustrated in Fig.~\ref{fig:pushblock_easy_cases}. With a longer high-level horizon, our method can form a more extended motion plan and succeeds repeatedly under more challenging initial layouts requiring larger directional corrections. Detailed per-task success rates for both Fruit25 and PushBlock are reported in Table~\ref{tab:realworld_per_task_results}.

\begin{table}[htbp]
\centering
\small
\setlength{\tabcolsep}{3.5pt}
\renewcommand{\arraystretch}{1.08}
\caption{
Motion smoothness and execution continuity on Fruit25 trials.
Values are reported as median [Q1, Q3].
Higher SPARC and lower jerk and stop ratio indicate smoother and more continuous execution.
}
\label{tab:fruit25_smoothness}
\resizebox{\linewidth}{!}{
\begin{tabular}{l c c c c c c}
\toprule
\textbf{Method}
& \textbf{T-SPARC $\uparrow$}
& \textbf{R-SPARC $\uparrow$}
& \textbf{T-Jerk $\downarrow$}
& \textbf{R-Jerk $\downarrow$}
& \textbf{Stop (\%) $\downarrow$}\\
\midrule

StarVLA-GR00T
& -5.480 [-5.817, -5.127]
& -10.499 [-11.932, -9.440]
& 9.599 [9.408, 9.788]
& 10.026 [9.833, 10.216]
& 12.537 [10.575, 15.102]\\

StarVLA-$\pi$
& -6.799 [-7.266, -6.233]
& -12.134 [-13.370, -11.135]
& 9.939 [9.681, 10.212]
& 10.355 [10.091, 10.599]
& 26.066 [23.566, 28.767]\\

\rowcolor{gray!12}
Ours
& \textbf{-5.151 [-5.398, -4.887]}
& \textbf{-9.992 [-11.099, -8.959]}
& \textbf{9.484 [9.300, 9.644]}
& \textbf{9.909 [9.733, 10.091]}
& \textbf{11.274 [8.974, 13.607]}\\

\bottomrule
\end{tabular}
}
\end{table}

\subsection{Motion Smoothness and Execution Continuity}
\label{sec:realworld_smoothness}
We further analyze the motion smoothness and execution continuity of the real robot using independently recorded high-rate feedback. Since PushBlock contains only 10 evaluation trials per method, we restrict this analysis to Fruit25. Both successful and failed trials are included. Table~\ref{tab:fruit25_smoothness} reports the median and interquartile range (IQR) over these trials. We use median-based statistics to reduce sensitivity to occasional outlier trajectories and long execution pauses.

We use three complementary measures of motion quality. \textbf{Translational SPARC} and \textbf{Rotational SPARC} are spectral arc-length measures computed from the magnitudes of the end-effector linear and angular velocities~\citep{balasubramanian2015analysis}, respectively; values closer to zero indicate smoother motion. \textbf{Translational} and \textbf{Rotational dimensionless jerk} measure the temporal variation of translational and rotational motion after normalization by trajectory duration and motion magnitude~\citep{hogan2009sensitivity}; we report the same $\log_{10}$ dimensionless jerk cost used throughout our analysis, where lower values indicate smoother motion. \textbf{Stop ratio} is the fraction of the active-motion interval for which the end-effector linear velocity falls below \(0.01\,\mathrm{m/s}\), with lower values indicating fewer execution pauses. 

As shown in Table~\ref{tab:fruit25_smoothness}, our hierarchical policy produces smoother and more continuous real-world execution, with fewer pauses.

%% file: ref.bib
@inproceedings{wei2026libra,
  title={Libra-VLA: Achieving Learning Equilibrium via Asynchronous Coarse-to-Fine Dual-System},
  author={Wei, Yifei and Zhong, Linqing and Liu, Yi and Lu, Yuxiang and He, Xindong and Yao, Maoqing and Ren, Guanghui},
  booktitle={Proceedings of the 64th Annual Meeting of the Association for Computational Linguistics (Volume 1: Long Papers)},
  pages={39799--39815},
  year={2026}
}

@misc{kim2024openvla,
      title={OpenVLA: An Open-Source Vision-Language-Action Model}, 
      author={Moo Jin Kim and Karl Pertsch and Siddharth Karamcheti and Ted Xiao and Ashwin Balakrishna and Suraj Nair and Rafael Rafailov and Ethan Foster and Grace Lam and Pannag Sanketi and Quan Vuong and Thomas Kollar and Benjamin Burchfiel and Russ Tedrake and Dorsa Sadigh and Sergey Levine and Percy Liang and Chelsea Finn},
      year={2024},
      eprint={2406.09246},
      archivePrefix={arXiv},
      primaryClass={cs.RO},
      url={https://arxiv.org/abs/2406.09246}, 
}

@article{pertsch2025fast,
  title={Fast: Efficient action tokenization for vision-language-action models},
  author={Pertsch, Karl and Stachowicz, Kyle and Ichter, Brian and Driess, Danny and Nair, Suraj and Vuong, Quan and Mees, Oier and Finn, Chelsea and Levine, Sergey},
  journal={arXiv preprint arXiv:2501.09747},
  year={2025}
}

@article{chi2025diffusion,
  title={Diffusion policy: Visuomotor policy learning via action diffusion},
  author={Chi, Cheng and Xu, Zhenjia and Feng, Siyuan and Cousineau, Eric and Du, Yilun and Burchfiel, Benjamin and Tedrake, Russ and Song, Shuran},
  journal={The International Journal of Robotics Research},
  volume={44},
  number={10-11},
  pages={1684--1704},
  year={2025},
  publisher={Sage Publications Sage UK: London, England}
}

@article{bjorck2025gr00t,
  title={Gr00t n1: An open foundation model for generalist humanoid robots},
  author={Bjorck, Johan and Casta{\~n}eda, Fernando and Cherniadev, Nikita and Da, Xingye and Ding, Runyu and Fan, Linxi and Fang, Yu and Fox, Dieter and Hu, Fengyuan and Huang, Spencer and others},
  journal={arXiv preprint arXiv:2503.14734},
  year={2025}
}

@article{black2024pi0,
  title={{$\pi_0$}: A Vision-Language-Action Flow Model for General Robot Control},
  author={Black, Kevin and Brown, Noah and Driess, Danny and Esmail, Adnan and Equi, Michael and Finn, Chelsea and Fusai, Niccolo and Groom, Lachy and Hausman, Karol and Ichter, Brian and others},
  journal={arXiv preprint arXiv:2410.24164},
  year={2024}
}

@article{intelligence2025pi05,
  title={{$\pi_{0.5}$}: a Vision-Language-Action Model with Open-World Generalization},
  author={{Physical Intelligence} and Black, Kevin and Brown, Noah and Darpinian, James and Dhabalia, Karan and Driess, Danny and Esmail, Adnan and Equi, Michael and Finn, Chelsea and Fusai, Niccolo and others},
  journal={arXiv preprint arXiv:2504.16054},
  year={2025}
}

@article{guo2026brain,
  title={A brain-inspired embodied intelligence for fluid and fast reflexive robotics control},
  author={Guo, Weiyu and Zhang, He and Li, Pengteng and Cai, Tiefu and Chen, Ziyang and Guo, Yandong and He, Xiao and Yang, Yongkui and Sun, Ying and Xiong, Hui},
  journal={arXiv preprint arXiv:2601.14628},
  year={2026}
}

@inproceedings{zhang2025action,
 author = {Zhang, Thomas T. and Pfrommer, Daniel and Pan, Chaoyi and Matni, Nikolai and Simchowitz, Max},
 booktitle = {International Conference on Learning Representations},
 editor = {C. Vondrick and B. Hariharan and C. Raffel and L. Pinto and D. Yang and A. Faust},
 pages = {126689--126742},
 title = {Action Chunking and Data Augmentation Yield Exponential Improvements in Behavior Cloning for Continuous Spaces},
 url = {https://proceedings.iclr.cc/paper_files/paper/2026/file/cd96cb9a239c37b39dbf34f3f5a4c56f-Paper-Conference.pdf},
 volume = {2026},
 year = {2026}
}

@article{niu2026t,
  title={T-rex: Tactile-reactive dexterous manipulation},
  author={Niu, Dantong and Liu, Zhuoyang and Wang, Zekai and Shao, Boning and Yin, Zhao-Heng and Pai, Anirudh and Sharma, Yuvan and Saravalle, Stefano and Zheng, Ruijie and Wang, Jing and others},
  journal={arXiv preprint arXiv:2606.17055},
  year={2026}
}

@INPROCEEDINGS{Zhao-RSS-23, 
    AUTHOR    = {Tony Z. Zhao AND Vikash Kumar AND Sergey Levine AND Chelsea Finn}, 
    TITLE     = {{Learning Fine-Grained Bimanual Manipulation with Low-Cost Hardware}}, 
    BOOKTITLE = {Proceedings of Robotics: Science and Systems}, 
    YEAR      = {2023}, 
    ADDRESS   = {Daegu, Republic of Korea}, 
    MONTH     = {July}, 
    DOI       = {10.15607/RSS.2023.XIX.016} 
}

@inproceedings{li2025hamster,
  title={Hamster: Hierarchical action models for open-world robot manipulation},
  author={Li, Yi and Deng, Yuquan and Zhang, Jesse and Jang, Joel and Memmel, Marius and Garrett, Caelan and Ramos, Fabio and Fox, Dieter and Li, Anqi and Gupta, Abhishek and others},
  booktitle={International Conference on Learning Representations},
  volume={2025},
  pages={24040--24068},
  year={2025}
}

@article{Fang2024MOKAOR,
  title={Moka: Open-world robotic manipulation through mark-based visual prompting},
  author={Liu, Fangchen and Fang, Kuan and Abbeel, Pieter and Levine, Sergey},
  journal={arXiv preprint arXiv:2403.03174},
  year={2024}
}

@article{hu2023look,
  title={Look before you leap: Unveiling the power of gpt-4v in robotic vision-language planning},
  author={Hu, Yingdong and Lin, Fanqi and Zhang, Tong and Yi, Li and Gao, Yang},
  journal={arXiv preprint arXiv:2311.17842},
  year={2023}
}

@InProceedings{shi2025hi,
  title = 	 {Hi Robot: Open-Ended Instruction Following with Hierarchical Vision-Language-Action Models},
  author =       {Shi, Lucy Xiaoyang and Ichter, Brian and Equi, Michael Robert and Ke, Liyiming and Pertsch, Karl and Vuong, Quan and Tanner, James and Walling, Anna and Wang, Haohuan and Fusai, Niccolo and Li-Bell, Adrian and Driess, Danny and Groom, Lachy and Levine, Sergey and Finn, Chelsea},
  booktitle = 	 {Proceedings of the 42nd International Conference on Machine Learning},
  pages = 	 {54919--54933},
  year = 	 {2025},
  editor = 	 {Singh, Aarti and Fazel, Maryam and Hsu, Daniel and Lacoste-Julien, Simon and Berkenkamp, Felix and Maharaj, Tegan and Wagstaff, Kiri and Zhu, Jerry},
  volume = 	 {267},
  series = 	 {Proceedings of Machine Learning Research},
  month = 	 {13--19 Jul},
  publisher =    {PMLR},
  url = 	 {https://proceedings.mlr.press/v267/shi25d.html}
}

@article{cui2025openhelix,
  title={Openhelix: A short survey, empirical analysis, and open-source dual-system vla model for robotic manipulation},
  author={Cui, Can and Ding, Pengxiang and Song, Wenxuan and Bai, Shuanghao and Tong, Xinyang and Ge, Zirui and Suo, Runze and Zhou, Wanqi and Liu, Yang and Jia, Bofang and others},
  journal={arXiv preprint arXiv:2505.03912},
  year={2025}
}

@misc{ye2026starvla,
      title={StarVLA: A Lego-like Codebase for Vision-Language-Action Model Developing}, 
      author={{StarVLA Community}},
      year={2026},
      eprint={2604.05014},
      archivePrefix={arXiv},
      primaryClass={cs.RO},
      url={https://arxiv.org/abs/2604.05014}, 
}

@article{liu2023libero,
  title={Libero: Benchmarking knowledge transfer for lifelong robot learning},
  author={Liu, Bo and Zhu, Yifeng and Gao, Chongkai and Feng, Yihao and Liu, Qiang and Zhu, Yuke and Stone, Peter},
  journal={Advances in Neural Information Processing Systems},
  volume={36},
  pages={44776--44791},
  year={2023}
}

@article{falck2025fourierspaceperspectivediffusion,
  title={A fourier space perspective on diffusion models},
  author={Falck, Fabian and Pandeva, Teodora and Zahirnia, Kiarash and Lawrence, Rachel and Turner, Richard and Meeds, Edward and Zazo, Javier and Karmalkar, Sushrut},
  journal={arXiv preprint arXiv:2505.11278},
  year={2025}
}

@article{balaji2022ediffi,
  title={ediff-i: Text-to-image diffusion models with an ensemble of expert denoisers},
  author={Balaji, Yogesh and Nah, Seungjun and Huang, Xun and Vahdat, Arash and Song, Jiaming and Zhang, Qinsheng and Kreis, Karsten and Aittala, Miika and Aila, Timo and Laine, Samuli and others},
  journal={arXiv preprint arXiv:2211.01324},
  year={2022}
}

@inproceedings{zhang2024improving,
  title={Improving training efficiency of diffusion models via multi-stage framework and tailored multi-decoder architecture},
  author={Zhang, Huijie and Lu, Yifu and Alkhouri, Ismail and Ravishankar, Saiprasad and Song, Dogyoon and Qu, Qing},
  booktitle={2024 IEEE/CVF Conference on Computer Vision and Pattern Recognition (CVPR)},
  pages={7372--7381},
  year={2024},
  organization={IEEE}
}

@inproceedings{reuss2025efficient,
  title={Efficient diffusion transformer policies with mixture of expert denoisers for multitask learning},
  author={Reuss, Moritz and Pari, Jyothish and Agrawal, Pulkit and Lioutikov, Rudolf},
  booktitle={International Conference on Learning Representations},
  volume={2025},
  pages={17247--17275},
  year={2025}
}

@inproceedings{hoeg2025fast,
  title={Fast policy synthesis with variable noise diffusion models},
  author={H{\o}eg, Sigmund H and Du, Yilun and Egeland, Olav},
  booktitle={2025 IEEE International Conference on Robotics and Automation (ICRA)},
  pages={4821--4828},
  year={2025},
  organization={IEEE}
}

@article{chen2025responsive,
  title={Responsive noise-relaying diffusion policy: Responsive and efficient visuomotor control},
  author={Chen, Zhuoqun and Yuan, Xiu and Mu, Tongzhou and Su, Hao},
  journal={arXiv preprint arXiv:2502.12724},
  year={2025}
}

@inproceedings{
li2026step,
title={{STEP}: Warm-Started Visuomotor Policies with Spatiotemporal Consistency Prediction},
author={Jinhao Li and Yuxuan Cong and Yingqiao Wang and Hao Xia and Shan Huang and Yijia Zhang and Ningyi Xu and Guohao Dai},
booktitle={Forty-third International Conference on Machine Learning},
year={2026},
url={https://openreview.net/forum?id=nWtjaAmug8}
}

@article{du2026cf,
  title={CF-VLA: Efficient Coarse-to-Fine Action Generation for Vision-Language-Action Policies},
  author={Du, Fan and Yan, Feng and Wu, Jianxiong and Xu, Xinrun and Zhang, Weiye and Wang, Weinong and Guo, Yu and Qian, Bin and He, Zhihai and Wang, Fei and others},
  journal={arXiv preprint arXiv:2604.24622},
  year={2026}
}

@article{chen2026streamvla,
  title={StreamVLA: Breaking the Reason-Act Cycle via Completion-State Gating},
  author={Chen, Tongqing and Wu, Hang and Wang, Jiasen and Li, Xiaotao and Fang, Lu},
  journal={arXiv preprint arXiv:2602.01100},
  year={2026}
}

@article{sun2026unifs,
  title={UniFS: Unified Fast-to-Slow Hierarchical Architecture for Vision-Language-Action Models},
  author={Sun, Lin and Guan, Zhiwei and Wang, Conglin and Chen, Zihong and Yu, Jianhai and Li, Zongsheng and He, Boyong and Sun, Tao and Cao, Jiale and Liu, Lige},
  journal={arXiv preprint arXiv:2606.22794},
  year={2026}
}

@article{wu2026coarse,
  title={Coarse-to-Control: Action-Token Planning for Vision-Language-Action Models},
  author={Wu, Jinhao and Zhang, Shiduo and Liu, Yicheng and Yu, Xiaopeng and Li, Sixian and Wang, Siyin and Zhao, Hang and Huo, Jing and Gao, Yang and Gong, Jingjing and others},
  journal={arXiv preprint arXiv:2606.07107},
  year={2026}
}

@article{du2025himoe,
  title={Himoe-vla: Hierarchical mixture-of-experts for generalist vision-language-action policies},
  author={Du, Zhiying and Liu, Bei and Liang, Yaobo and Shen, Yichao and Cao, Haidong and Zheng, Xiangyu and Feng, Zhiyuan and Wu, Zuxuan and Yang, Jiaolong and Jiang, Yu-Gang},
  journal={arXiv preprint arXiv:2512.05693},
  year={2025}
}

@misc{kim2025finetuningvisionlanguageactionmodelsoptimizing,
      title={Fine-Tuning Vision-Language-Action Models: Optimizing Speed and Success}, 
      author={Moo Jin Kim and Chelsea Finn and Percy Liang},
      year={2025},
      eprint={2502.19645},
      archivePrefix={arXiv},
      primaryClass={cs.RO},
      url={https://arxiv.org/abs/2502.19645}, 
}

@inproceedings{fei2026libero,
  title={Libero-plus: A progressive robustness benchmark for visual-language-action models},
  author={Fei, Senyu and Wang, Siyin and Shi, Junhao and Dai, Zihao and Cai, Jikun and Qian, Pengfang and Ji, Li and He, Xinzhe and Zhang, Shiduo and Fei, Zhaoye and others},
  booktitle={Proceedings of the IEEE/CVF Conference on Computer Vision and Pattern Recognition},
  pages={38574--38583},
  year={2026}
}

@article{balasubramanian2015analysis,
  title={On the analysis of movement smoothness},
  author={Balasubramanian, Sivakumar and Melendez-Calderon, Alejandro and Roby-Brami, Agnes and Burdet, Etienne},
  journal={Journal of neuroengineering and rehabilitation},
  volume={12},
  number={1},
  pages={112},
  year={2015},
  publisher={Springer}
}

@article{hogan2009sensitivity,
  title={Sensitivity of smoothness measures to movement duration, amplitude, and arrests},
  author={Hogan, Neville and Sternad, Dagmar},
  journal={Journal of motor behavior},
  volume={41},
  number={6},
  pages={529--534},
  year={2009},
  publisher={Taylor \& Francis}
}
